%% file: main.tex
\documentclass[10pt, a4paper]{article}

\usepackage[final]{lrec2026} 

\usepackage{times}
\usepackage{latexsym}
\usepackage{comment}

\usepackage[T1]{fontenc}

\usepackage[utf8]{inputenc}

\usepackage{microtype}

\usepackage{inconsolata}
\usepackage{subfig}

\usepackage{amsmath, amsthm}
\usepackage{graphicx}
\usepackage{xcolor}
\usepackage{color, soul}
\usepackage{pgfplots}
\usepackage{pgfplotstable}
\usepackage{subfig}
\usepackage{booktabs}
\usepackage{footnote}
\usepackage{todonotes}
\makesavenoteenv{tabular}
\makesavenoteenv{table}

\definecolor{darkishred}{HTML}{a90308}

\newcommand{\darkgreen}

\definecolor{bronze}{rgb}{0.8, 0.5, 0.2}

\definecolor{coolblack}{rgb}{0.0, 0.18, 0.39}

\definecolor{cornflowerblue}{rgb}{0.39, 0.58, 0.93}

\theoremstyle{definition}

\title{The Moralization Corpus: Frame-Based Annotation and Analysis of Moralizing Speech Acts across Diverse Text Genres}

\name{Maria Becker, Mirko Sommer, Lars Tapken, Yi Wan Teh, Bruno Brocai} 

\address{Department of German Linguistics, Heidelberg University \\ Department of Computational Linguistics, Heidelberg University \\
         \{maria.becker, bruno.brocai\}@gs.uni-heidelberg.de\\
         \{sommer, tapken, teh\}@cl.uni-heidelberg.de\\}

\abstract{
Moralizations -- arguments that invoke moral values to justify demands or positions -- are a yet underexplored form of persuasive communication. We present the Moralization Corpus, a novel multi-genre dataset designed to analyze how moral values are strategically used in argumentative discourse. Moralizations 
are pragmatically complex and often implicit, posing significant challenges for both human annotators and NLP systems. We develop a frame-based annotation scheme that captures the constitutive elements  of moralizations -- moral values, demands, and discourse protagonists -- and apply it to a diverse set of German texts, including political debates, news articles, and online discussions. The corpus enables fine-grained analysis of moralizing language across communicative formats and domains.
We further evaluate several large language models (LLMs) under varied prompting conditions for the task of moralization detection and moralization component extraction and compare it to human annotations in order to investigate the challenges of automatic and manual analysis of moralizations. Results show that detailed prompt instructions have a greater effect than few-shot or explanation-based prompting, and that moralization remains a highly subjective and context-sensitive task. We release all data, annotation guidelines, and code to foster future interdisciplinary research on moral discourse and moral reasoning in NLP.
 \\ \newline  }

\begin{document}

\maketitleabstract

\input{1introduction-new.tex}

\input{2RelatedWork}

\input{3data.tex}

\input{4stats}

\input{5experiments.tex}

\input{6eval}

\input{7conclusion.tex}

\section*{Limitations}
While the Moralization Corpus constitutes a unique resource for studying moralizing speech acts across genres, several limitations remain. First, both the automatic and manual evaluation of moral value and protagonist classification can be further improved. The current evaluation setup provides initial insights into model behavior, but more fine-grained semantic and boundary-sensitive measures are needed to better capture the nuanced character of moral references and role assignments. Second, although our experiments included configurations that prompted models to verbalize explanations, we have not yet systematically analyzed the content, coherence, or validity of these explanations. Future work will therefore include a dedicated investigation into how explanation quality correlates with model accuracy and human interpretability.

Third, no model fine-tuning has yet been performed on our dataset. Since the annotation scheme introduces new task-specific concepts such as moral frames and pragmatic roles, fine-tuned models might substantially improve the detection and classification of moralizations. A fourth limitation concerns the contextual scope of our instances: the current dataset is based on five-sentence snippets, which, while sufficient for local pragmatic analysis, may not always capture the full discursive context in which moralizations unfold. Expanding the contextual window or including paragraph-level annotations will thus be an important step toward a more comprehensive understanding of moral reasoning in discourse.

Furthermore, although our dataset is multilingual in structure, detailed annotations have so far only been carried out for the German data, therefore language-specific lexical, grammatical, and pragmatic conventions may influence how moralization is expressed, potentially limiting the direct transferability of the results to other languages. Future work will extend the annotation framework to English, French, and Italian, enabling cross-linguistic comparisons and broader generalization. Finally, due to the high complexity and time intensity of the task, parallel double annotation was conducted only for selected subsets rather than for the entire corpus. While our multi-step adjudication ensured consistency and reliability, a fully parallel annotation process would further strengthen inter-annotator agreement and improve the overall robustness of the dataset.

\section*{Ethics Statement}
The Moralization Corpus was constructed using publicly available texts and copyright-compliant material (e.g., parliamentary debates, news articles, online discussions) and does not include private or sensitive data. Given the inherently normative character of moral discourse, annotators were trained to focus on linguistic and pragmatic aspects rather than moral evaluation or agreement with the content. We acknowledge that subjectivity is an integral part of moral interpretation; our multi-step annotation protocol and adjudication procedures were designed to minimize bias while preserving interpretive diversity.
We further emphasize that the goal of this research is analytical and descriptive, not prescriptive: the dataset and models are not intended for moral judgment or behavioral prediction, but to support interdisciplinary research on the intersection of communication, argumentation, and moral framing.

\section*{Acknowledgements}
We would like to thank all student assistants involved in the annotation of the Moralization Corpus for their dedication and careful work. We are also grateful to our colleagues for their insightful discussions on annotation methodology and moral discourse. This research was supported by institutional funding and by the interdisciplinary research project \textit{Moralisierungen in der Wissenschaftskommunikation (MoWiKo)}, funded by the Federal Ministry of Research, Technology and Space (BMFTR), Germany. In addition, the authors gratefully acknowledge the computing time provided on the high-performance computer HoreKa by the National High-Performance Computing Center at KIT (NHR@KIT). This center is jointly supported by the Federal Ministry of Education and Research and the Ministry of Science, Research and the Arts of Baden-Württemberg, as part of the National High-Performance Computing (NHR) joint funding program (\url{https://www.nhr-verein.de/en/our-partners}). HoreKa is partly funded by the German Research Foundation (DFG).

\section{Bibliographical References}\label{sec:reference}

\bibliographystyle{lrec2026_natbib}
\bibliography{anthology,custom}

\appendix
\input{appendix.tex}

\end{document}

%% file: 1introduction-new.tex
\section{Introduction}
\label{sec:introduction}

Recently, an increasing number of studies at the interface of Natural Language Processing (NLP) and Computational Social Science (CSS) have addressed the task of modeling morality in text, reflecting the growing interest in exploring moral phenomena through computational means. Most of this work has either focused on predicting moral values from text 
(e.g., \citealp{doi:10.1080/19331681.2013.826613, 10.1145/2858036.2858423, Diakopoulos2014IdentifyingAA}), or on analyzing moral biases in large language models (LLMs) (\citealp{schramowski2022large, hendrycks2021aligning, haemmerl-etal-2023-speaking, Jiang2021DelphiTM, fraser-etal-2022-moral}, among others). However, the pragmatic patterns of moralizations -- that is, how moral values are strategically used in argumentative contexts to justify demands or stances -- have not yet been systematically modeled in NLP research.

We understand moralizations as persuasive strategies in which moral values are invoked to describe controversial topics and to demand specific actions or judgments \cite{felder2022diskurs, becker2023moral}. Three examples appear in Table \ref{tab:exa}. In moralizing practices, vocabulary associated with moral values (e.g., \textit{freedom, justice, security, inequality}) serves to reinforce a demand by linking it to widely shared moral norms \cite{haidt-etal-2009, graham-etal-2013}. For instance, in the sentence \textit{We should introduce a refugee cap in order to ensure the safety of Germans}, the term \textit{safety} functions as a moral justification for a political demand. As the examples in Table \ref{tab:exa} illustrate, moralizations can take many forms and occur in a broad range of contexts -- political, social, religious, and even scientific -- beyond explicitly populist or manipulative discourse. That said, our goal is not to assess moralizations in terms of being ``good'' or ``bad,'' but rather to examine their linguistic realization and discourse functions.

\begin{table}[]
\scalebox{0.85}{
    \begin{tabular}{p{8.7cm}}
    \hline
         (1) \textit{We should all stop eating meat because it causes \textcolor{teal}{\textbf{unnecessary suffering to animals}}}.
          \\ 
         \\
       (2) \textit{Women still earn less than men, even though \textcolor{teal}{\textbf{equality between men and women}} is enshrined in the Basic Law.} \\
       \\
       (3) \textit{Immigrants are taking jobs from \textcolor{teal}{\textbf{hardworking citizens}} and undermining \textcolor{teal}{\textbf{our values}}}. \\
       \hline
    \end{tabular}}
    \caption{Examples of moralizations from our dataset (moral phrases in bold) illustrate how moral values support explicit (Ex. 1) or implicit (Ex. 2–3) demands and occur in both populistic (Ex. 3) and non-populistic (Ex. 1–2) contexts.}
    \label{tab:exa}
\end{table}

While previous computational studies have primarily modeled morality through simplified categorical frameworks (e.g., one moral value per tweet or sentence), the complexity and heterogeneity of moralizations as a pragmatic phenomenon call for a more nuanced and structured approach. In this paper, we therefore propose a novel annotation framework for modeling moralization frames in text, which captures the interplay between moral values, demands, and discourse protagonists across multiple text genres. 
Annotating moral values and demands links values to concrete prescriptions, revealing how moral arguments serve as legitimation strategies. Identifying protagonists as moral agents, beneficiaries, or culprits uncovers the social dynamics behind such arguments. 
Thus, being able to identify and analyze moralizations in different text genres provides a valuable foundation for interesting research e.g. in linguistics, social and political sciences.\footnote{Similar to \citet{rehbein-etal-2025-moral}, our focus is not on aligning LLMs with human values or investigating moral biases in LLMs, but instead to use NLP approaches to analyze moralizations in different texts, thereby contributing to research on value-based reasoning.}

Our framework is designed to operationalize moralizations in a way that makes their linguistic and pragmatic properties empirically accessible. It is holistic, in that it integrates moral, rhetorical, and argumentative dimensions within a unified frame structure; and flexible, in that it can be applied to various languages, genres, and segment sizes. The annotation process involves iterative refinement, combining qualitative and quantitative validation steps to ensure coherence and reliability.

By applying this framework to a German dataset comprising political debates, media reports, and online discussions, we show that certain types of moralizations and discourse roles become analytically visible only through our multidimensional annotation. In addition, our annotation studies highlight the inherent subjectivity of the task, revealing how differing conceptual understandings of moralization influence annotation consistency. We also probe several LLMs for their ability to detect moralizations automatically, providing both a feasibility study and a detailed error analysis that sheds light on which linguistic and contextual factors are decisive for successful moralization detection.

Our \textbf{contributions} are threefold: (1) We propose a novel, frame-based annotation framework for moralizations that captures moral values, demands, and protagonists and allows for a fine-grained analysis of moralizations; (2) We apply and refine this framework across diverse genres of German texts, demonstrating its analytical potential for investigating moral rhetoric and framing, e.g. in political and social discourse; and (3) We conduct and compare several manual annotation and LLM-based detection experiments together with extensive evaluations to explore the challenges of identifying moralizations. All resources developed in this work -- including the annotated dataset, annotation manual, and code -- are released publicly to support further research in this area.\footnote{\url{https://github.com/GS-Uni-Heidelberg/Paper-TheMoralizationCorpus}} In sum, our goal is to provide a methodological foundation for analyzing how moral rhetoric operates across discourses -- a foundation that, we argue, enables new insights into moral communication that previous modeling approaches could not reveal.

The remainder of the paper is structured as follows: \S\ref{sec:relatedwork} provides an overview of prior work on (computational) modeling of morality, then \S \ref{sec:data} and \ref{sec:stats} introduce our annotation framework and dataset. \S  \ref{sec:experiments} and \S \ref{sec:eval} report our experiments and evaluation for moralization detection, and \S \ref{sec:conclusion} discusses implications and future directions. 

%% file: 2RelatedWork.tex
\section{Related Work}
\label{sec:relatedwork}

Morality has been extensively studied both within NLP and in other disciplines; however, the specific phenomenon of moralization and its computational modeling have so far received little attention.

\begin{figure*}[]
\begin{centering}
\includegraphics[width=0.8
\textwidth]{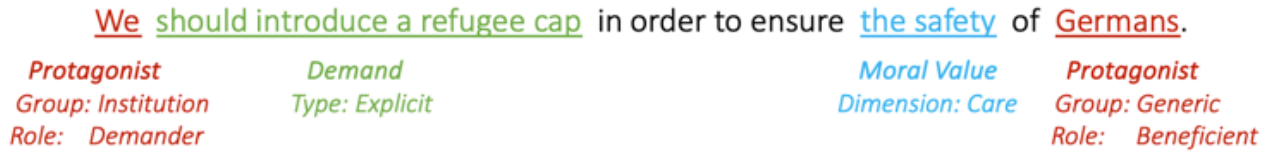}
\caption{Fully annotated example of a moralization frame, labeled with the demand, the supporting moral value and the protagonists (all translations from German in this paper are by the authors).} 
\label{fig:example-anno}
\end{centering}
\end{figure*}

\paragraph{Morality in Computational Social Science (CSS).} As pointed out by \citet{reinig-etal-2024-survey}, many approaches computationally model morality in order to investigate research questions from the political or social sciences. These studies predict moral attitudes or sentiments from newspaper or social media text, e.g. on discourses about abortion policies \cite{10.1145/2858036.2858423}, vaccine campaigns \cite{10021123} or climate change   \cite{Diakopoulos2014IdentifyingAA}. For all studies of morality in the field of CSS, Twitter is by far the most prominent empirical basis (see \citeauthor{reinig-etal-2024-survey} \citeyear{reinig-etal-2024-survey}).
In contrast, our dataset captures moralizations across heterogeneous genres and communicative formats, including more subtle contexts such as non-fiction.

\paragraph{Interdisciplinary Perspectives.}
Outside NLP, moralization has been studied in linguistics \citep{felder2022diskurs, becker2023moral}, communication studies \citep{kampf2016political}, psychology \citep{rhee2019and}, and political science \citep{mooijman2018moralization}. These works emphasize moralization as a persuasive strategy -- a phenomenon largely unexplored computationally. Our study addresses this gap by analyzing the potential of computationally modeling moralizations.

\paragraph{Morality and Argumentation.}
We examine moral values in argumentative contexts. Work at this intersection (e.g., \citealp{kobbe-etal-2020-exploring, kiesel-etal-2023-semeval}) shows that moral values contribute to argumentative quality and persuasion. The goal of the SemEval shared task ValueEval'23 
\citep{kiesel-etal-2023-semeval} is to explore if it is possible to automatically uncover the values on which arguments draw; and \citet{Landowska_Budzynska_Zhang_2024} label political texts preannotated for argument structures with moral foundations in order to explain the strategies in the use of moral arguments.

\paragraph{Frame-Based Approaches.}
Few studies model morality in relation to entities or events. \citet{roy-etal-2022-towards, roy-etal-2021-identifying} define morality frames as the moral foundation(s) invoked by a text, along with the sentiment toward mentioned entities. 
Similarly, \citet{lei-etal-2024-emona} and \citet{zhang-etal-2024-moka} combine moral foundations with entity and event information in order to learn morality-relevant text representations. The approach by \citet{rehbein-etal-2025-moral}, most related to ours, provides fine-grained moral frame annotations including moral frame types, moral foundations, and narrative roles in parliamentary debates. We extend this line toward broader genres and pragmatic functions. 

\paragraph{Subjectivity in Annotations.} Moral labeling is inherently subjective \citep{falk2025mining, chochlakis2024larger}. Recent work proposes multi-annotator and justification-based methods \citep{weber2024varierr, alvarez2024moral}. We follow this direction with a multi-step annotation pipeline designed to capture and justify subjective variation (see §\ref{sec:data}).


%% file: 3data.tex
\section{Dataset and Annotation}
\label{sec:data}

Our dataset of moralizing text passages was created in four steps: (1) development of a dictionary of morality-indicating words; (2) retrieval of text snippets from large corpora and web sources based on the dictionary entries; (3) creation of an annotation scheme capturing key components of moralizations; and (4) a multi-step annotation process ensuring data quality and consistency.

\paragraph{Dictionary Creation.}
To identify moralizing passages, we developed \textsc{DiMi}, a multilingual dictionary of morality-indicating words. Starting from a manually curated German seed list of 130 words (e.g., \textit{freedom, fairness, guilt}) \citep{felder2022diskurs}, we expanded it using co-occurrence profiles from the CCDB database \citep{belica2011semantische}. 
After manual cleaning, the dictionary comprised 3,000 entries, which we automatically translated into English, French, and Italian and manually verified.\footnote{Unlike existing moral dictionaries, our resource includes not only explicitly moral terms but also contextually moralizing words such as \textit{guise}. Importantly, the lexicon itself does not distinguish between explicit and contextually moralizing terms; rather, this distinction is determined during annotation based on contextual use.}

\paragraph{Data Collection.}
Next, we used \textsc{DiMi} to query large corpora and online sources for text passages containing at least one dictionary entry. For each language (German, English, French, Italian), we retrieved 2,000 five-sentence snippets from seven genres: letters to the editor, interviews (both from various newspapers), parliamentary debates (plenary minutes), commentaries (opinion articles), court reports (newspaper articles about legal cases), Wikipedia discussions (where users discuss how to improve an article), and non-fiction books (on history, parenting, cultural studies, etc.). The German data were drawn from \textsc{DeReKo} \cite{ids2022dereko}, while other languages were collected from publicly available web texts. Each dataset was split into training (70\%), development (15\%), and test (15\%) sets, balanced across genres.

\paragraph{Category Development.}
Our annotation captures three interrelated layers, designed to capture the key pragmalinguistic features of moralizations and to enable consistent corpus-based analysis across different genres and contexts. Figure~\ref{fig:example-anno} illustrates a fully annotated example.

(1) \textbf{Moral values} (phrase level), mapped to the six Moral Foundations \textsc{Care/Harm, Fairness/Cheating, Loyalty/Betrayal, Authority/Subversion, Purity/Degradation}, and \textsc{Liberty/Oppression} according to the Moral Foundations Theory (MFT) \citep{graham-etal-2009, graham-etal-2013}.\footnote{The MFT assumes that moral reasoning is driven by a set of intuitive emotional responses, or “gut feelings”,  that underlie and rationalize moral judgment.} Multi-label assignments are allowed, 
and multiple values in the same instance are annotated separately;
(2) \textbf{Demands} (clause or sentence level), here annotators mark all explicit demands (e.g. \textit{We should all stop eating meat}) in the texts. For implicit demands, annotators rephrase the claim in a simple sentence (e.g. \textit{Women still earn less than men} is explicated as \textit{Women should be paid equally}); and 
(3) \textbf{Protagonists} (phrase level), labeled by (a) group type: \textsc{Individuals} (e.g. \textit{Angela Merkel}), \textsc{Generic} (references to humans, such as \textit{the people, men and women}), \textsc{Institutions/Organizations} (e.g. \textit{the democrats, the stakeholders}), and \textsc{Social Groups} (e.g. \textit{parents, homeless people}); and (b) discourse role (role within the moralization): person who is moralizing (\textsc{Demander}), target of the demand (\textsc{Adressee}), person who would benefit (\textsc{Beneficiary}) or being disadvantaged (\textsc{Maleficiary}) from the demand.

Together, these layers form a \textbf{moralization frame}, which we define as the text span that links moral values, demands, and protagonists. These elements are constitutive of moralizations and encompass all central components necessary for the systematic analysis and interpretation of moralizations in discourse.\footnote{We acknowledge that retrieval may yield incomplete frames; this limitation is addressed in our evaluation, and future work will focus on methods for extracting and segmenting complete frames.} 

\paragraph{Annotation Procedure.}
The annotation of moralizations is not only complex but also inherently subjective. To address this, we designed a multi-step annotation procedure that captures nuanced judgments without compromising the operationalization required for computational modeling. As outlined above, annotations (and analysis) were conducted only for the German dataset; extending the annotation framework to the English, French, and Italian data is currently in progress. The annotation proceeded in six stages:

\begin{table}[]
\scalebox{0.85}{
    \begin{tabular}{p{4.2cm}|p{4.2cm}}
    \hline
         (a) \textit{The mayor remains silent on topics such as child poverty, while publicly championing prestige projects. Millions are being wasted on the mistakes of the senators. It is time that politicians, too, are held accountable.} \textrightarrow \textcolor{teal}{\textbf{Moralization}} &
       (b) \textit{Researchers collected data on child poverty and other broader social issues, and how politicians respond to them in public discourse. The material was analyzed to identify recurring themes across different text types.}
       \textrightarrow \textcolor{purple} {\textbf{No Moralization/ NVI}}
       \\
       \hline
    \end{tabular}}
    \caption{Passages retrieved with \textsc{DiMi}, where (a) constitutes a moralization while (b) neutrally describes a research activity about child poverty.}
    \label{tab:exa2}
\end{table}

(1) \textbf{Identification}: The mere occurrence of a moral word or phrase does not in itself constitute a moralization (see Table \ref{tab:exa2}). In fact, we observe that in many contexts, moral terms may also appear in neutral or reportive texts without carrying any persuasive or argumentative intention (referred to as \textit{Neutral Value-Referring Instances}, NVIs). Distinguishing between NVIs and moralizations was thus a crucial first step in our annotations.
We therefore prepared a check list for moralizations together with positive and negative examples. Then, each retrieved text passage is annotated independently by two annotators (binary classification), both with a background in linguistics, yielding a high agreement of 0.71 (Cohen's Kappa). Passages with disagreement are adjudicated by an expert annotator (one of the authors). 

(2) \textbf{Pilot phase}: We then conducted an initial exercise using a subset of 200 moralizations to train six annotators (all with a background in linguistics) for the task of moral value component detection and classification (values, demands and protagonists). Besides annotator training, the goal was to optimize the annotation manual by identifying sources of annotation variance, problematic categories, and ambiguous examples. We iteratively refined the manual through additional examples and special-case rules, before proceeding to the annotation of the full dataset. 

(3) \textbf{Full annotation}: All instances that have been identified as moralizations in the identification step were then distributed across the six trained annotators who labeled moral values, demands, and protagonists based on our codebook using the INCEpTION platform \citep{tubiblio106270}.

(4) \textbf{Review}: Each file was secondarily reviewed by another annotator, allowing for corrections and additions, and generating a further set of discussion items.\footnote{This approach has been inspired by similar approaches such as \citet{weber2024varierr} or \citet{becker-etal-2024-detecting}. Since moralization feature annotation is highly time-intensive, full parallel annotation of the dataset was not feasible.}  Open questions or ambiguous cases were discussed and resolved with the team. 

(5) \textbf{Re-annotation of NVIs}: While the review focused on moralizing instances, some entries labeled as NVI qualify, upon closer inspection, as moralizations. These NVIs were re-reviewed by three expert annotators, limited to the test set due to the time-intensive process.\footnote{A refined test set suffices for the prompting experiments in this paper, automated re-annotation of the full dataset is underway for future finetuning experiments.} Instances confirmed as moralizations were retroactively annotated with all moralization components, resulting in 278 additional annotated moralization instances.

(6) \textbf{Formal validation}: Final consistency checks ensured that the formal annotation rules had been respected (e.g., that each moralization contains a demand or spans were marked correctly). We correct and supplement the data accordingly and remove irrecoverable instances. 

The resulting Moralization Corpus provides a rich, pragmatically grounded resource for studying moralizations across genres and serves as a benchmark for computational modeling of moralizing discourses.

%% file: 4stats.tex
\section{Data Statistics}

\label{sec:stats}


This section provides an overview of the dataset and its main characteristics. Illustrative examples are provided throughout the text; additional examples are listed in \S\ref{sec:appendix-exa}.

\paragraph{Overview.} The final dataset contains 11,503 instances with an average length of 83 tokens, evenly distributed across seven genres (for detailed statistics, see Table \ref{tab:stats2}). The proportion of moralizations varies, mainly due to the particular consideration of the test set within our multi-step annotation process (18\% in Dev and Train, and 45\% in Test). Across all genres, however, NVIs clearly outnumber moralizations, confirming that moral terms are often used descriptively rather than strategically. 

\paragraph{Moral Values.}
Within moralizations, an average of 1.6 moral values were annotated per instance, most often in Wikipedia discussions (1.7) and least frequently in online comments (1.4). Across the dataset, the \textsc{Care–Harm} pair dominates (\textsc{Care} 16\%, \textsc{Harm} 22\%), followed by \textsc{Fairness–Cheating} (16\%/13\%). \textsc{Authority} (3\%) and \textsc{Subversion} (2\%) are rare.
Genre variation aligns with communicative context (see Fig.  \ref{fig:1}): \textsc{Fairness–Cheating} is especially frequent in Wikipedia discussions (28\%/19\%), reflecting norms of equality and rule compliance in the meta-discourses about the editing of articles (see Example 1 and 2 in \S \ref{sec:appendix-exa}). 
In court reports, \textsc{Fairness} and \textsc{Liberty} as foundational principles of law prevail, while non-fiction books show more \textsc{Oppression} due to historical topics and war narratives.

\paragraph{Demands.}
Explicit and implicit demands are balanced overall (53\% vs. 47\%), but differ by genre (see Fig. \ref{fig:2}). Explicit demands dominate in parliamentary debates (66\%), while implicit ones prevail in commentaries (48\%) and letters to the editor (26\%), where addressees are diffuse publics rather than interlocutors. 
In such cases, moralizations serve less to direct action than to express evaluation or positioning, and accordingly, moral appeals often stay implicit, as in the following extract from a letter to the editor (see Example 3 in \S \ref{sec:appendix-exa} for the full text): \textit{It is outrageous how arrogantly our politicians ignore the needs of our children. From a political perspective, children do not pay off, but for society they certainly do.} 

\paragraph{Protagonists.}
Across all instances, the \textbf{roles} \textsc{beneficiaries} (which appear avg. 0.65 times within a moralization) and \textsc{addressees} (0.64) occur most frequently, followed by demanders (0.42); \textsc{maleficiaries} are rare (0.10) (see Fig.  \ref{fig:3} for the distribution across different genres). Moralizations therefore tend to emphasize positive outcomes rather than blame (see Example 4 and 5 in \S \ref{sec:appendix-exa}). 
In most cases, not all protagonist slots within the moralization frame are explicitly filled and must be inferred from context or world knowledge, as in the following example from a parliamentary debate where the beneficiary stays implicit: \textit{We need stricter laws to prevent racially motivated violence.} (see Example 6 in \S \ref{sec:appendix-exa} for the full instance.)
\textsc{Institutions} (32\%) and \textsc{social groups} (30\%) are the most frequent protagonist \textbf{group} types, followed by \textsc{individuals} (20\%) and  \textsc{generic human} references (15\%) (for the distribution across genres, see Fig. \ref{fig:4}). This suggests that moral demands often invoke collective actors, as group-level outcomes appear more socially relevant and persuasive than individual ones (see Example 7 in \S \ref{sec:appendix-exa}), 
consistent with previous findings on the social function of moralization \citep{becker2025diskursgrammatik}.

Typical \textbf{role–group configurations} (see Fig. \ref{fig:5})  show that individuals act as demanders, institutions as addressees, and social groups as beneficiaries. Moralizations thus reflect a characteristic pattern linking individual agency, institutional responsibility, and collective good.

These distributional patterns illustrate the analytical potential of the dataset for studying moral rhetoric and framing in discourse. By explicitly linking moral values, demands, and discourse protagonists, the annotation framework makes visible how moral arguments are structured and strategically deployed across communicative contexts. For instance, the prevalence of collective protagonists and beneficiary roles highlights the tendency of moralizations to frame issues in terms of collective welfare, while the balance between explicit and implicit demands shows that moral claims are often conveyed indirectly through evaluative or argumentative framing. These observations demonstrate how the corpus can be used not only to detect moral language but also to analyze the rhetorical mechanisms through which moral values shape argumentative discourse.

%% file: 5experiments.tex
\section{Experiments}

\label{sec:experiments}

Building on these observations, we then explore how computational methods can support the analysis of moralizations in text. To this end, we evaluate several LLMs using different prompt designs to assess their ability to detect moralizations. The central objective, however, is a systematic comparison of model and human judgments in light of the inherent subjectivity of the task, which is why our evaluation focuses in particular on both human–model and human–human agreement.

\paragraph{Prompt Engineering.}
Our prompts\footnote{ \url{https://github.com/GS-Uni-Heidelberg/Paper-TheMoralizationCorpus/tree/main/prompts}} are derived from our annotation manual and follow its structure\footnote{ \url{https://github.com/GS-Uni-Heidelberg/Paper-TheMoralizationCorpus/tree/main/manual}}. Each defines moralization by three criteria: (1) the presence of moral values, (2) an explicit or implicit demand, and (3) an argumentative link between both. The prompts further guide the extraction and classification of moral values and protagonists. Output is generated in a standardized JSON format including all components and a short explanation: Moral phrases and their classification according to MFT, extracted or reconstructed demands, protagonists together with their assigned roles and group affiliations, the binary decision on whether the passage constitutes a moralization, and a short explanatory rationale. The underlying chain of thought requires the model to proceed step by step: first identifying the core components of moralizations, 
and then deciding 
whether the text qualifies as a moralization.

Prompt versions were refined through iterative evaluations. Key adjustments improved recall and precision: (a) clearer definition of positive and negative values, (b) stricter rules for identifying implicit demands, (c) stronger emphasis on the argumentative link, (d) explicit description of NVIs, and (e) integration of borderline cases into the examples.

\paragraph{Prompt Configurations.}
We experimented with seven configurations varying in level of detail, reasoning requirement, and example inclusion:
(1) \texttt{basic-0shot}: minimal instruction;
(2) \texttt{cot-0shot}: stepwise reasoning with detailed instructions but without examples;
(3) \texttt{cot-10shot}: same, plus ten examples;
(4) \texttt{cot-explain-0shot}: here the model must explicitly verbalize its reasoning (an explanatory step), no examples;
(5) \texttt{cot-explain-10shot}: explanation plus examples;
(6) \texttt{manual-0shot}: a detailed configuration based on our annotation manual; and 
(7) \texttt{manual-explain-0shot}: manual plus explanatory step.
Each of the five models was tested with all seven configurations, resulting in 35 outputs per instance.

\paragraph{Models.}
We tested five state-of-the-art instruction-tuned LLMs differing in architecture, scale, and context window, allowing
us to assess prompt performance across models
with varying memory and reasoning capacities: 
LLaMA-4-Scout-17B-16E-Instruct (109B), C4AI-Command-a-03-2025 (111B), Mistral-Small-3.2-24B-Instruct-2506 (24B), GPT-5-mini-2025-08-07, and Claude-3.5-Haiku-20241022. 
 Model parameters are displayed in Table \ref{tab:mod}.

%% file: 6eval.tex
\section{Evaluation and Analysis}

\label{sec:eval}

\begin{figure*}[t]
\centering
\includegraphics[width=0.9\linewidth]{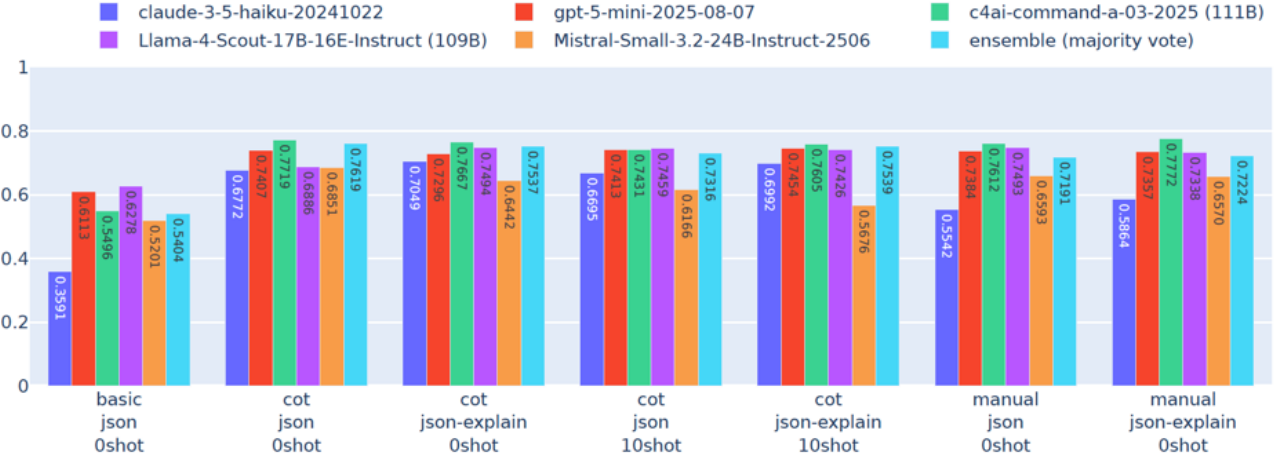}
\caption{Binary moralization classification across models and prompting conditions (macro F1, test set).}
\label{fig:results}
\end{figure*}

We evaluated annotation quality and model performance for (a) binary moralization detection and (b) component extraction and classification (values, demands, and protagonists). 

\subsection{Binary Moralization Classification}

\label{sec:eval-binary}

Fig. \ref{fig:results} summarizes results on the test set across all prompting conditions for the binary moralization classification task (detailed classification results for each model, including precision and recall scores, are displayed in the Appendix, 
Table \ref{tab:a} to \ref{tab:f}). Performance differences are generally small; detailed prompts yield the most consistent gains above the basic version, underscoring that a clear, structured definition of moralization is crucial.
Among models, Cohere attains the best F1 scores, followed by the ensemble model (majority vote of all five models); Claude and Mistral lag behind. Notably, few-shot examples and forced explanations do not consistently improve performance, suggesting that moralization requires deeper pragmatic reasoning than these techniques capture.
Error analysis shows a precision-recall trade-off: detailed prompts reduce false positives ($\uparrow$ precision, $\downarrow$ recall), while example-enriched prompts reduce false negatives ($\uparrow$ recall, $\downarrow$ precision). In practical applications, the choice therefore depends on whether recall (e.g. for monitoring or detection systems) or precision (e.g. for analytical research tasks) is prioritized. 

\subsection{Moralization Component Detection and Classification}
\label{sec:eval-moral-components}
\paragraph{Moral values \& Protagonists.}
Automatically evaluating whether a model has identified and classified all relevant moral values and protagonists within a moralization is particularly challenging, since precise boundary detection and overlapping labels, among others, limit automatic agreement with gold annotations.

We evaluate moral value and protagonist spans with the SemEval-2013 NER-style setup \cite{segura-bedmar-etal-2013-semeval} using strict and partial matching (see \S \ref{sec:appendix-met} for details). 
\textbf{Moral values} achieve low F1 scores (strict $\leq$ 0.20; partial up to 0.22; see Fig.~\ref{fig:eval1}, \ref{fig:eval2}, \ref{fig:eval3}, and \ref{fig:eval4}), indicating difficulties with span boundaries and context-sensitive, often implicit value expressions. Detailed prompts (opposed to basic descriptions) boost performance most, while examples and explanation generation yield small but consistent gains.

\textbf{Protagonists} perform higher (strict F1 of 0.20–0.28; +0.03–0.05 pp under partial, see Fig. \ref{fig:eval5}, \ref{fig:eval6}, \ref{fig:eval7}, \ref{fig:eval8}, \ref{fig:eval9}, and \ref{fig:eval10} for details).  Here, detailed prompts and examples help recall while precision remains limited, reflecting overgeneration and multi-label ambiguity. 

Across models, Cohere and Mistral perform best for values, while GPT leads for protagonists.

Methodologically, results are constrained by our selective annotation scheme (focus on morally relevant values/actors) and the task's subjectivity, so metrics should be read as indicative rather than definitive.  Nevertheless, the consistent relative ranking across models and strategies provides a first indication of system behavior and points to directions for targeted fine-tuning and nuanced evaluation.

\paragraph{Demands.}
Next, we evaluate the models' ability to extract or generate demand formulations. 
We employ a combination of BLEU \citep{papineni-etal-2002-bleu}, ROUGE \citep{lin-2004-rouge}, both measuring lexical overlap; and BERTScore \citep{zhang2020bertscore} which leverages embeddings to estimate semantic similarity.

The results 
show only minimal variation between models and prompt configurations. BERTScore ranges from 73 (Mistral,  \texttt{manual-explain}) to 78 (Cohere,  \texttt{cot-0shot}). As expected, explicit demands yield substantially higher performance (up to 82) compared to implicit ones (maximum 75). Overlap metrics are lower, as expected for free text generation tasks. 

Given the limitations of reference-based evaluation in this setting (cf.  \citealp{becker-etal-2021-reconstructing}), 
we additionally conducted a manual evaluation of extracted (in case of explicit) or generated (in case of implicit) demands. Therefore, two team annotators assessed a subset of 73 moralizations (see \S\ref{sec:agreement} for the selection process). For each instance, the demands extracted or generated by the three best-performing configurations (see \S\ref{sec:eval-binary}) of each of the five models were collected. In total, 1,095 demands were rated on a five-point Likert scale for semantic correctness, i.e., how accurately the demand conveyed the intended moral claim. 
Agreement was substantial (Cohen’s Kappa = 0.63), and remaining disagreements were resolved by an expert adjudicator. 

Results appear in Fig. \ref{fig:app-demand} and show that models achieved high average ratings of 4.27, indicating that generated demands largely captured the intended moral argumentation. Differences across configurations were minimal, and no notable differences appeared between implicit and explicit demands. Cohere performed best (up to 4.6), while Mistral scored lowest (3.8).

Overall, these results confirm that moralization is a complex linguistic and conceptual phenomenon challenging for automated detection and interpretation. Although detailed prompts and strong models improve performance slightly, overall scores remain moderate, highlighting the need for both model adaptation and refined evaluation methods.

\subsection{Agreement between Models and Humans}\label{sec:agreement}

Next, we take a closer look at these results to explore moralization patterns across genres and to interpret common deviations between human and model-based annotation -- with the ultimate goal of informing linguistic and social-scientific analysis of moralizations in discourse.

\paragraph{Annotation Setup.} To assess moralization detection challenges for both humans and LLMs, we selected a genre-balanced subset of 150 test instances (Test-150), consisting of 73 moralizations and 77 NVIs. The subset was subsequently annotated by five annotators with varying levels of expertise: Expert 1 (project lead, >2 years), Experts 2 \& 3 (doctoral researchers, >1 year), and Student Assistant 1 \& 2 (few months of experience). This setup enabled analysis of how project familiarity -- and thus understanding of our definitions -- affects annotation consistency. All annotators received the same detailed prompts as the models and made binary yes/no moralization judgments (on the instance level).


\paragraph{Findings.} Moralization rates increased with project familiarity: Students labeled 23–24\% as moralizations, while Experts averaged 38\%. This suggests that familiarity broadens detection, as everyday notions are narrower than our operational definition. LLMs labeled 59\%, indicating a more liberal classification tendency (see \S \ref{sec:eval-agr}, \S \ref{sec:eval-ind}).

\begin{figure*}[t]
\begin{centering}
\includegraphics[width=0.85
\textwidth]{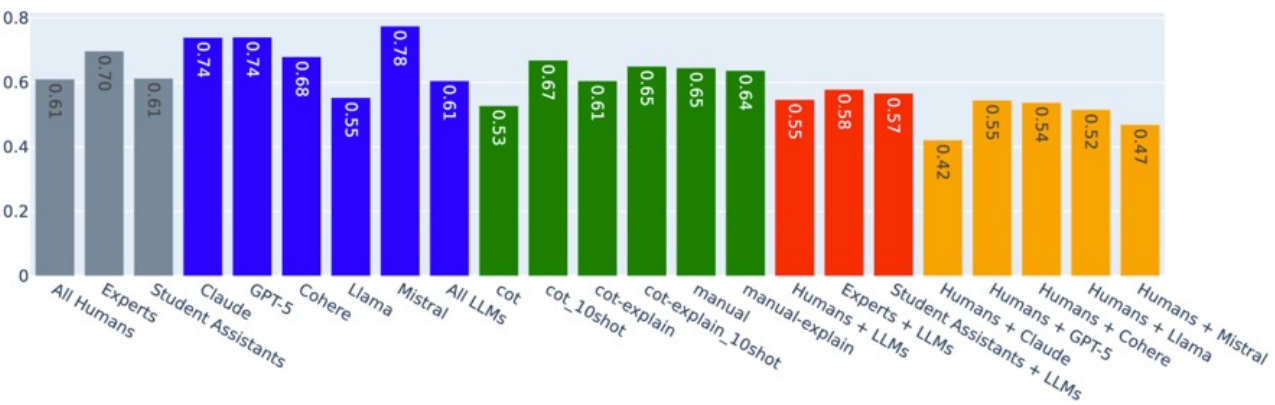}
\caption{Mean PABAK Scores for different comparisons of agreement between and within humans and models. Fleiss’ scores (on avg. 1–2 pp lower) follow precisely the same tendencies, cf. Fig. \ref{fig:kappa}.} 
\label{fig:pabak}
\end{centering}
\end{figure*}

Next, human-human, model-model, and human-model agreement were compared using Fleiss' Kappa and PABAK (which adjusts Kappa for prevalence and bias). Results (see Fig. \ref{fig:pabak}, for Fleiss' Kappa which follows the same tendencies cf. Fig. \ref{fig:kappa})
show that experts agree more with each other than students, and student labels align more with LLMs than experts do. Models agree with each other to a degree comparable to human–human agreement, but expert–expert consistency is highest. Interestingly, the weakest models in binary classification (Claude, Mistral) show the highest consistency across prompts. Explanation prompts increase agreement between models, suggesting that they foster more consistent interpretations, even if they do not improve overall predictive performance, as shown in \S \ref{sec:eval-binary}.

\subsection{Analysis of (Dis)Agreement}
\label{sec:eval-agr}

To understand both the limits of model-based moralization detection and the characteristic patterns in human versus model reasoning, we analyzed cases of agreement and disagreement within Test-150 and collected the following statistics, focusing on the most clearly classified cases (for a tabular overview, see Table \ref{tab:agreement}): How often, and in which cases, do (1) ...all annotators agree (5/5) or at least 80\% ($\geq80$\% 4/5)? (2) ...do all models/configurations agree (35/35) or at least 80 80\% ($\geq80$\% 29/35)? (3) ...do all models and humans agree (40 identical decisions), or at least 80\% of each group coincide? (4) ...do models and humans diverge fundamentally (i.e., $\geq80$\%80\% of models vs. $\geq80$\%80\% of humans make opposite decisions)?

Results 
show that models agree more on moralizations (43\%) than on NVIs (24\%); humans show the opposite pattern (22\% vs. 57\%). Instances with total agreement among all humans and models are displayed in Example 8 and 9 in  \S \ref{sec:appendix-exa}.
In total, only twelve cases can be identified in which human and model decisions fundamentally diverge, see Example 10 in \S\ref{sec:appendix-exa} for an illustrative instance.

Overall, the results of our (dis-)agreement analysis suggest that models rely strongly on surface cues, whereas humans capture more subtle or implicit cases. To explore this observation further, we examine lexical cues in the data in the next subsection.


\subsection{Linguistic Indicators of Moralization}
\label{sec:eval-ind}

To test the hypothesis that specific text features function as the primary drivers for model (and human) decisions, a selected range of linguistic indicators was examined which, according to \citet{becker2025diskursgrammatik}, may serve as cues for the classification of moralizations. Examples for each category appear in Table \ref{tab:exa22}.

\paragraph{Categories.} 
(1) Text genre: Moralizations may be easier to identify in certain genres (e.g., opinionated texts); (2) Moral vocabulary: A high frequency of moral words ($\geq5$ per instance) may impel both humans and models to label a text as a moralization; (3) Explicit demands: are likely easier to recognize than implicit ones; (4) Modal verbs: As markers of deontic modality, they often co-occur with explicit demands and thus can serve as cues for moralization; 
(5) Subjunctive mood: Since moralizations often describe future scenarios and subsequent actions, subjunctive forms can work as signals; and (6) Instance length: Very short fragments might lack sufficient context for classification and are more likely labeled as NVIs.

Focusing on cases with $\ge$80\% model–human agreement, we found no genre effects, but other indicators showed trends largely confirming our hypotheses: We find high moral-term density in 90\% of agreed moralizations vs. only 25\% of agreed NVIs; modal verbs in 71\% of moralizations vs. 22\% in NVIs; subjunctive is also more frequent in moralizations (23\% vs. 8\%); and explicit demands occur in 81\% of unanimously moralizations. Finally, the prevalence of short snippets among agreed NVIs points to the need for dataset refinement, as instance length stems from data extraction artifacts rather than linguistic content. A systematic overview of the results is provided in Table \ref{tab:indi}.

\subsection{Deviation Analysis (Models Only)}
\label{sec:eval-var}

Finally, we compared model predictions with the human majority vote for Test-150 to identify typical divergences. The analysis distinguishes between False Positives (FP -- model predictions of moralization where the human majority vote is NVI) and False Negatives (FN -- model predictions of NVI where the human majority vote is moralization).\footnote{For reasons of space, we summarize the key findings below; further details on the annotation procedure and results can be found in the Appendix, \S \ref{sec:details-deviation}.}

Our analysis and annotations reveal three main sources for \textbf{FPs}. (i) Neutral uses of moral vocabulary: Many models label passages as moralizing whenever words such as \textit{duty, moral,
responsibility, right,} or \textit{wrong} appear, even
when used descriptively or within quotations, see Example 11 in \S\ref{sec:appendix-exa}; (ii) missing context: In some cases, the surrounding context is missing, making it impossible
to determine whether moralization is present (Example 12); and (iii) borderline cases: Since moralization detection
is a subjective task, certain instances can
reasonably be interpreted either way. These
ambiguous cases are those where moralization
could plausibly be argued even though
the majority human label is NVI (Example 13). Better-performing models (LLaMA, Cohere, GPT) show more borderline FPs, suggesting finer sensitivity, while weaker models (Mistral, Claude) more often mislabel neutral passages. Dominant cause for \textbf{FNs} are missed demands, especially implicit ones. Less frequent causes are missed moral terms and missed value-demand link across sentences, as well as figurative language, irony, and negated demands.

\begin{table}[]
\scalebox{0.85}{
    \begin{tabular}{p{8.7cm}}
    \hline
         (4) \textit{It is not enough to simply stand up and say: We reject \textcolor{teal}{\textbf{all forms of violence}} with \textcolor{teal}{\textbf{disgust}} and \textcolor{teal}{\textbf{indignation}}. We
must not only show that we do not tolerate \textcolor{teal}{\textbf{racism}} and \textcolor{teal}{\textbf{violence}}, but also convey the \textcolor{teal}{\textbf{democratic val-
ues}} by which we want to convince our children. Therefore,
I believe that under no circumstances should we
restrict \textcolor{teal}{\textbf{civil rights}}.} (Parliamentary debates)
           \\ 
         \\
         (5) \textit{Attac \textcolor{teal}{\textbf{should}} advocate for a Global Marshall Plan for developing countries. Poverty reduction \textcolor{teal}{\textbf{can}} only succeed if infrastructure problems are addressed: expanding education systems, enforcing women’s rights, and ensuring access to energy and water. In addition, international institutions such as the IMF, WTO, and World Bank \textcolor{teal}{\textbf{must}} be democratized.}
         (Letters to the editor)
          \\ 
         \\
       (6) \textit{Politically, however, resistance from the left \textcolor{teal}{\textbf{would}} not only jeopardize the currency reform but also the constitutional basis for the Solidarity Foundation. What is needed now is a truly Swiss-style compromise.} (Interviews) \\
       \hline
    \end{tabular}}
    \caption{Examples for lexical cues of moralizations (in bold): density of moral words (4), modal verbs (5), and subjunctives (6).}
    \label{tab:exa22}
\end{table}

\paragraph{Summary.}
Overall, the analyses show that moralization detection is inherently subjective, with models relying on linguistic cues and humans -- especially experts -- capturing more implicit moralizations. Our findings highlight the challenges of evaluating and modeling nuanced moral reasoning, shaped by subjectivity, context, and linguistic variability.

%% file: 7conclusion.tex
\section{Conclusion}
\label{sec:conclusion}

In this paper, we introduced the Moralization Corpus, a novel, frame-based resource for analyzing how moral values are strategically employed in argumentative discourse. The empirical patterns observed in the dataset -- including the distribution of moral values, demands, and discourse roles -- illustrate how the annotation framework enables systematic analysis of moral framing strategies in discourse. Our framework operationalizes moralization as the interplay between moral values, demands, and discourse participants, allowing for a fine-grained analysis of how moral rhetoric functions across genres. The annotation procedure and resulting data shed light on the pragmatics of moralizing communication, demonstrating that moralizations often rely on implicit reasoning and situated inference rather than (only) overt moral vocabulary, and that many moralizing demands are realized implicitly and require contextual interpretation. Experimental results with several LLMs show that detailed task definitions are essential for reliable moralization detection, whereas few-shot examples and explanation generation do not consistently improve performance. Human-model comparisons further reveal that both groups face similar challenges -- particularly regarding implicitness, subjectivity, and the pragmatic boundaries of moral speech acts.
Taken together, this work provides an empirical and methodological foundation for future research on moral communication, argumentation, and persuasion. Beyond linguistic and social-scientific applications, our results also inform computational modeling of complex, subjective, and pragmatically grounded language phenomena.

%% file: appendix.tex
\section{Appendix}
\label{sec:appendix}

\subsection{Dataset Statistics and Visualizations}
\label{sec:appendix-stats}

Table \ref{tab:stats2} provides additional statistical details on our dataset presented in \S \ref{sec:stats}.

\begin{table*}
\small
\centering
\refstepcounter{table}
\begin{tabular}{l|l|l|l|l}
\toprule
\textbf{Split}                 & \textbf{Instances} & \textbf{Percentage}   & \textbf{Proportion} & \textbf{AVG tokens}  \\ 
                 &  &    & \textbf{Moralizations} &  \\
\hline
test                           & 1,584               & 13.8          & 44.7                           & 82.7                 \\
Test-150                        & 150                & 1.3           & 48.5                           & 77.6                 \\
dev                            & 1,953               & 17            & 17.7                           & 84.0                   \\
train                          & 7,816               & 67.9          & 17.8                           & 83.2                 \\ 
\midrule
Court reports        & 1,337      & 11.6 & 12.1                           & 91.2        \\
Letters to the editor & 1,553      & 13.5 & 27.8                           & 84.7        \\
Parliamentary debates & 1,583      & 13.8 & 36.1                           & 74.4        \\
Wikipedia discussions & 1,659      & 14.4 & 7.3                            & 79.0          \\
Commentaries          & 1,807      & 15.7 & 33.6                           & 97.3        \\
Interviews            & 1,855      & 16.1 & 32.5                           & 62.7       \\
Non-fiction books     & 1,709      & 14.9 & 5.5                            & 95.1        \\ 
\midrule
\textbf{Total/Medium}          & \textbf{11,503}     & \textbf{100}  & \textbf{21.2}                           & \textbf{83.2}   \\
\bottomrule
\end{tabular}
\caption{Data split across train, development, and test sets and distribution of genres within the complete dataset, including the number of instances, proportion of the entire dataset, ratio of moralizations, and the average number of tokens per instance. Differences in the proportion of moralizations across the splits, particularly between the test set and the training and development sets, result from the refinement of annotations, as described in \S \ref{sec:stats}.}
\label{tab:stats2}
\end{table*}

\begin{figure*}
    \centering
    \includegraphics[width=1\linewidth]{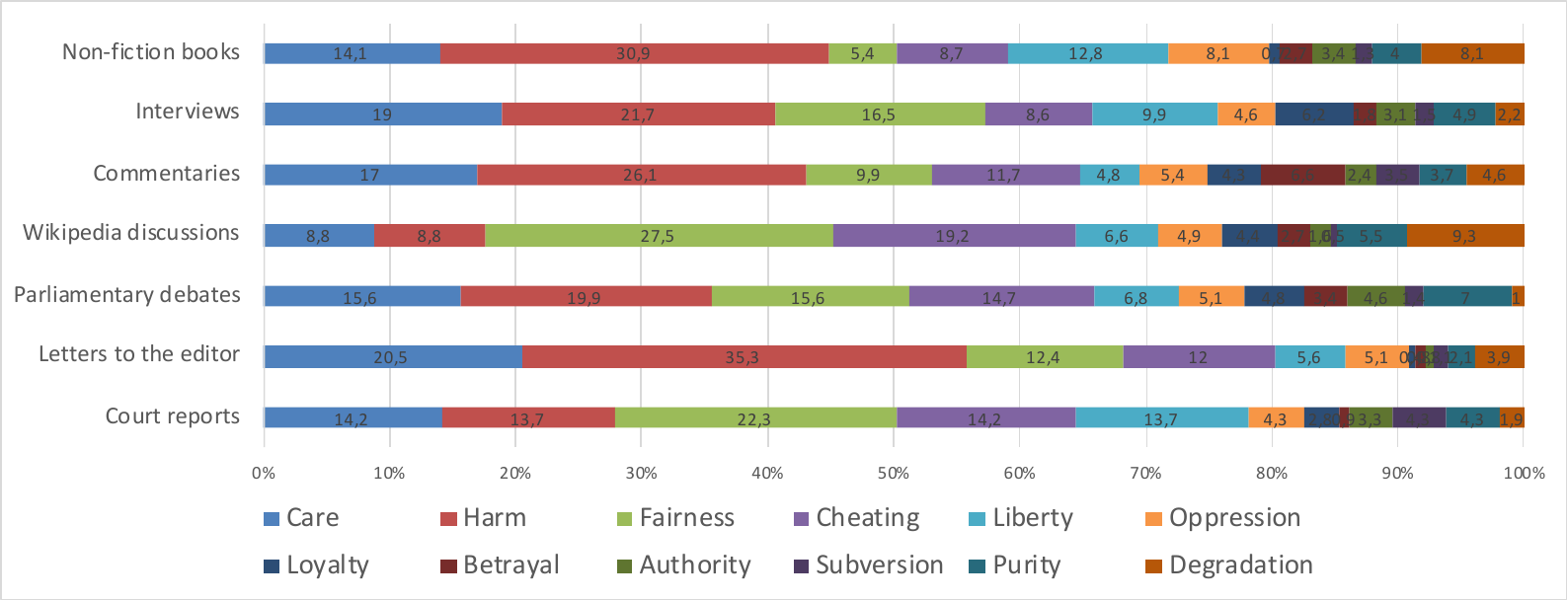}
    \caption{Distribution of moral values (according to MFT) across genres, numbers in percentage.}
    \label{fig:1}
\end{figure*}

\begin{figure*}
    \centering
    \includegraphics[width=0.65\linewidth]{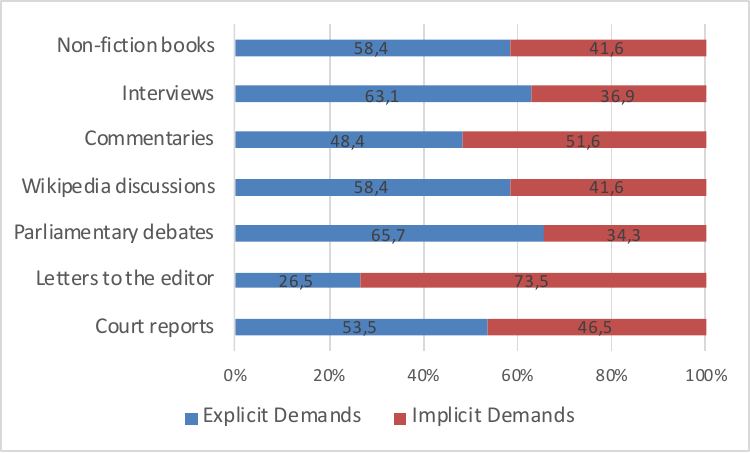}
    \caption{Distribution of explicit vs. implicit demands across genres, numbers in percentage.}
    \label{fig:2}
\end{figure*}

\begin{figure*}
    \centering
    \includegraphics[width=0.65\linewidth]{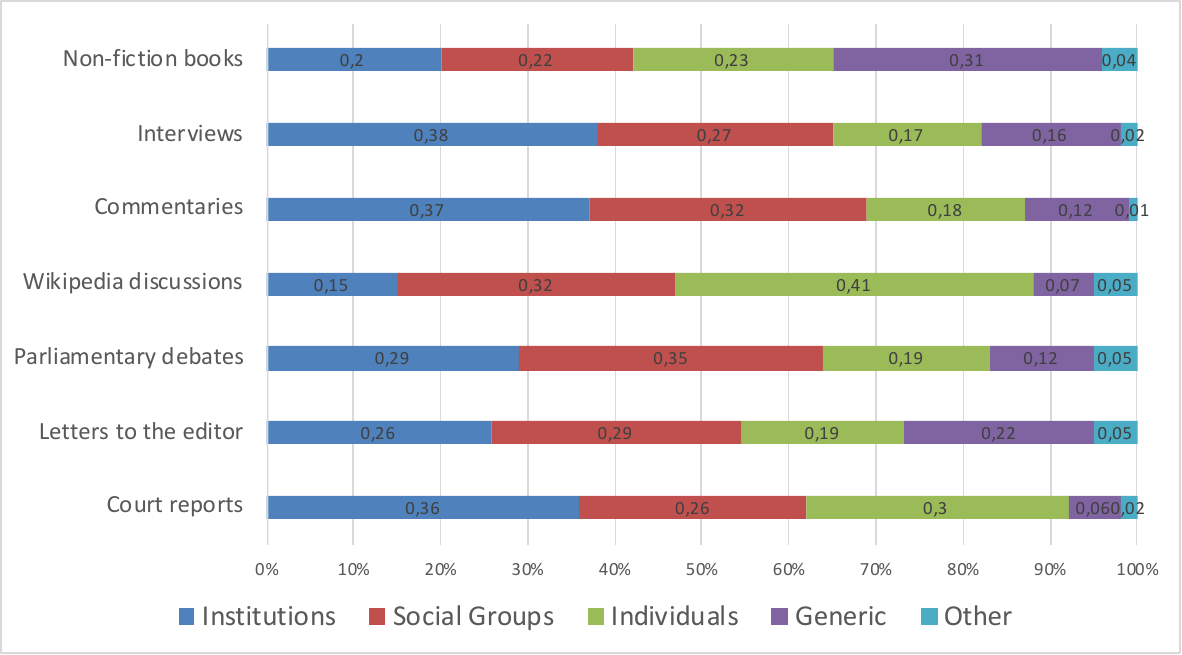}
    \caption{Distribution of groups across genres, numbers in percentage.}
    \label{fig:3}
\end{figure*}

\begin{figure*}
    \centering
    \includegraphics[width=0.65\linewidth]{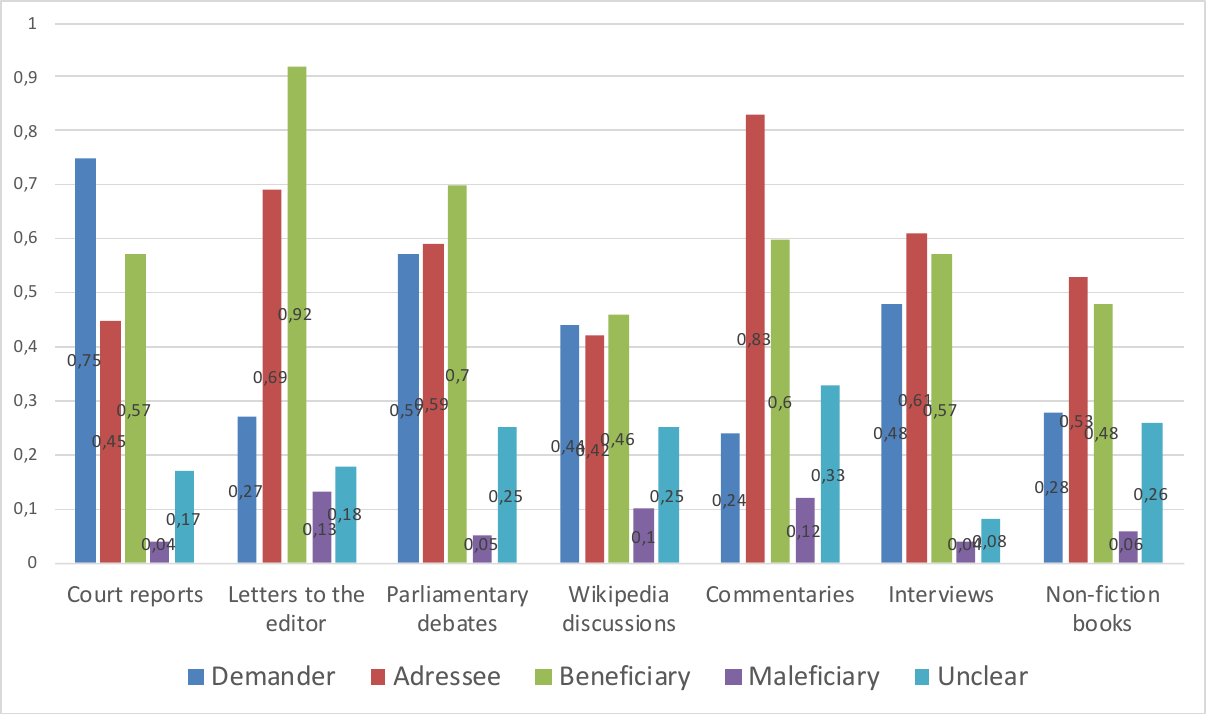}
    \caption{Distribution of moralizing roles across genres, numbers in total.}
    \label{fig:4}
\end{figure*}

\begin{figure*}
    \centering
    \includegraphics[width=0.65\linewidth]{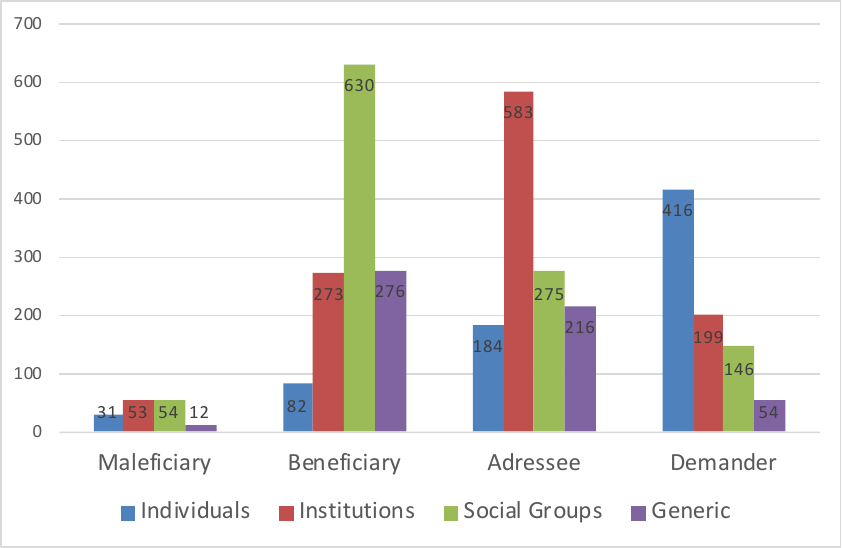}
    \caption{Co-occurrences of protagonist groups and roles, numbers in total.}
    \label{fig:5}
\end{figure*}

Visualizations and detailed information on the analysis of the data statistics (\S \ref{sec:stats}) are displayed in Fig. \ref{fig:1}, \ref{fig:2}, \ref{fig:3}, \ref{fig:4}, and \ref{fig:5}.

\label{sec:appendix-vis}

\subsection{Prompts}
\label{sec:appendix-pro}

Our prompts are divided into four sections. Part A defines the concept of moralization and specifies the three necessary criteria for its identification: (1) the presence of one or more moral values (based on the Moral Foundations Theory), (2) the presence of an explicit or implicit demand to act, refrain from acting, or adopt a stance, and (3) an argumentative link between the moral value and the demand. Part B outlines the extraction of protagonists, Part C their categorization into predefined social or institutional classes, and Part D the assignment of moralization roles (e.g., Demander, Addressee, Beneficiary). Examples illustrate each step, and the prompt concludes with a standardized JSON output format for computational processing. All prompt configurations can be found here: \url{https://github.com/GS-Uni-Heidelberg/Paper-TheMoralizationCorpus/tree/main/prompts}

\subsection{Models}
\label{sec:appendix-mod}
Table \ref{tab:mod} summarizes the statistics of the models used in our experiments (\S \ref{sec:experiments}).

\begin{table*}
\small
\centering
\begin{tabular}{l|l|l|l|l|l}
\toprule
             & llama & cohere & mistral & gpt-5 & claude  \\
             \hline
model size   & 109B~                                   & 111B                            & 24B                                          & not public                     & not public                          \\
context size & 10M                                     & 256K                            & 128K                                         & 400k                           & 200k                       \\        
\bottomrule
\end{tabular}
\caption{Statistics of the models used in our experiments.}
\label{tab:mod}
\end{table*}

\subsection{Results and Evaluation}
\label{sec:appendix-eval}

\subsubsection{Binary Moralization Classification}

Tables \ref{tab:a}, \ref{tab:b}, \ref{tab:c}, \ref{tab:d}, \ref{tab:e}, and \ref{tab:f} provide detailed results of our binary moralization classification experiments (see \S \ref{sec:eval-binary}). The tables report model-specific performance metrics, including precision and recall, allowing for a comprehensive comparison across different model architectures.

\begin{table}
\scriptsize
\centering
\begin{tabular}{l|l|l|l|l}
\toprule
Claude                              & acc    & pre    & rec    & f1      \\
\hline
basic\_0shot          & 0.3889 & 0.6635 & 0.5587 & 0.3568  \\
cot\_0shot            & 0.6788 & 0.7193 & 0.7500 & 0.6752  \\
cot\_10shot           & 0.6701 & 0.7200 & 0.7474 & 0.6675  \\
cot-explain\_0shot    & 0.7089 & 0.7313 & 0.7690 & 0.7029  \\
cot-explain\_10shot   & 0.7031 & 0.7300 & 0.7664 & 0.6978  \\
manual\_0shot         & 0.5527 & 0.6891 & 0.6711 & 0.5517  \\
manual-explain\_0shot & 0.5839 & 0.6965 & 0.6920 & 0.5839 \\
\bottomrule
\end{tabular}
\caption{Binary classification results for Claude.}
\label{tab:a}
\end{table}

\begin{table}
\scriptsize
\centering
\begin{tabular}{l|l|l|l|l}
\toprule
GPT-5                              & acc    & pre    & rec    & f1      \\
\hline
basic\_0shot          & 0.6094 & 0.7047 & 0.7099 & 0.6093  \\
cot\_0shot            & 0.7639 & 0.7324 & 0.7598 & 0.7398  \\
cot\_10shot           & 0.7645 & 0.7329 & 0.7602 & 0.7403  \\
cot-explain\_0shot    & 0.7506 & 0.7228 & 0.7533 & 0.7287  \\
cot-explain\_10shot   & 0.7679 & 0.7370 & 0.7653 & 0.7445  \\
manual\_0shot         & 0.7477 & 0.7425 & 0.7849 & 0.7364  \\
manual-explain\_0shot & 0.7448 & 0.7428 & 0.7854 & 0.7343 \\
\bottomrule
\end{tabular}
\caption{Binary classification results for GPT.}
\label{tab:b}
\end{table}

\begin{table}[t!]
\scriptsize
\centering
\begin{tabular}{l|l|l|l|l}
\toprule
Cohere                              & acc    & pre    & rec    & f1      \\
\hline
basic\_0shot          & 0.5486 & 0.6844 & 0.6666 & 0.5476  \\
cot\_0shot            & 0.7963 & 0.7634 & 0.7873 & 0.7718  \\
cot\_10shot           & 0.7604 & 0.7376 & 0.7736 & 0.7423  \\
cot-explain\_0shot    & 0.7917 & 0.7584 & 0.7819 & 0.7666  \\
cot-explain\_10shot   & 0.7812 & 0.7515 & 0.7823 & 0.7597  \\
manual\_0shot         & 0.7789 & 0.7535 & 0.7895 & 0.7605  \\
manual-explain\_0shot & 0.7980 & 0.7674 & 0.7970 & 0.7764 \\
\bottomrule
\end{tabular}
\caption{Binary classification results for Cohere.}
\label{tab:c}
\end{table}

\begin{table}[t!]
\scriptsize
\centering
\begin{tabular}{l|l|l|l|l}
\toprule
Llama                              & acc    & pre    & rec    & f1      \\
\hline
basic\_0shot          & 0.6262 & 0.6964 & 0.7125 & 0.6251  \\
cot\_0shot            & 0.7789 & 0.7738 & 0.6738 & 0.6915  \\
cot\_10shot           & 0.7656 & 0.7394 & 0.7731 & 0.7458  \\
cot-explain\_0shot    & 0.7980 & 0.7673 & 0.7397 & 0.7504  \\
cot-explain\_10shot   & 0.7656 & 0.7366 & 0.7673 & 0.7437  \\
manual\_0shot         & 0.7743 & 0.7418 & 0.7678 & 0.7497  \\
manual-explain\_0shot & 0.7506 & 0.7321 & 0.7696 & 0.7342 \\
\bottomrule
\end{tabular}
\caption{Binary classification results for Llama.}
\label{tab:d}
\end{table}

\begin{table}
\scriptsize
\centering
\begin{tabular}{l|l|l|l|l}
\toprule
Mistral                              & acc    & pre    & rec    & f1      \\
\hline
basic\_0shot          & 0.5208 & 0.6752 & 0.6465 & 0.5181  \\
cot\_0shot            & 0.6921 & 0.7097 & 0.7448 & 0.6849  \\
cot\_10shot           & 0.6157 & 0.6936 & 0.7060 & 0.6151  \\
cot-explain\_0shot    & 0.6441 & 0.7020 & 0.7239 & 0.6422  \\
cot-explain\_10shot   & 0.5666 & 0.6905 & 0.6795 & 0.5662  \\
manual\_0shot         & 0.6620 & 0.6992 & 0.7279 & 0.6578  \\
manual-explain\_0shot & 0.6586 & 0.7058 & 0.7322 & 0.6556 \\
\bottomrule
\end{tabular}
\caption{Binary classification results for Mistral.}
\label{tab:e}
\end{table}

\begin{table}
\scriptsize
\centering
\begin{tabular}{l|l|l|l|l}
\toprule
ensemble                              & acc    & pre    & rec    & f1      \\
\hline
basic\_0shot         & 0.5399 & 0.6911 & 0.6645 & 0.5380  \\
cot\_0shot            & 0.7824 & 0.7538 & 0.7863 & 0.7618  \\
cot\_10shot           & 0.7413 & 0.7398 & 0.7819 & 0.7308  \\
cot-explain\_0shot    & 0.7708 & 0.7485 & 0.7863 & 0.7537  \\
cot-explain\_10shot   & 0.7650 & 0.7559 & 0.8000 & 0.7532  \\
manual\_0shot         & 0.7280 & 0.7289 & 0.7691 & 0.7177  \\
manual-explain\_0shot & 0.7297 & 0.7365 & 0.7777 & 0.7211 \\
\bottomrule
\end{tabular}
\caption{Binary classification results for the Ensemble model (majority votes from all models).}
\label{tab:f}
\end{table}

\subsubsection{Metrics for Moral Value and Protagonist Evaluation}
\label{sec:appendix-met}

For the automatic evaluation of moral value detection and classification as well as protagonist detection and classification (see \S \ref{sec:eval-moral-components}), we adopted the evaluation framework for Named Entity Recognition (NER) models as defined in the SemEval 2013 - 9.1 task \cite{segura-bedmar-etal-2013-semeval}. Following this setup, we used the Python library nervaluate\footnote{\url{https://github.com/MantisAI/nervaluate}} to compute token-level and span-level agreement between system output and gold-standard annotations. The evaluation distinguishes between five outcome categories: COR (correct, exact match), INC (incorrect, mismatch between system and gold annotation), PAR (partial overlap), MIS (missed gold annotation), and SPU (spurious system output).

Based on these, possible items (POS) are defined as COR + INC + PAR + MIS (true positives and false negatives), and actual items (ACT) as COR + INC + PAR + SPU (true positives and false positives). We report both strict and partial match scores. Under strict matching, precision and recall are calculated as

\begin{align*}
\mathrm{Precision} &= COR / ACT = TP / (TP + FP)\\
\mathrm{Recall} &= COR / POS = TP / (TP + FN)
\end{align*}

Under partial matching, overlapping annotations are weighted by 0.5 to account for near matches:

\begin{align*}
\mathrm{Precision} &= (COR + 0.5 × PAR) / ACT\\
\mathrm{Recall} &= (COR + 0.5 × PAR) / POS
\end{align*}

Since our annotation scheme allows multiple MFT labels per moral value, we counted COR, INC, PAR, SPU occurrences for six virtue/vice pairs separately, then aggregated the counts across MFT class labels to compute precision and recall.

\subsubsection{Moral Value Evaluation}
\label{sec:appendix-mv}

\begin{figure*}
    \centering
    \includegraphics[width=0.8\linewidth]{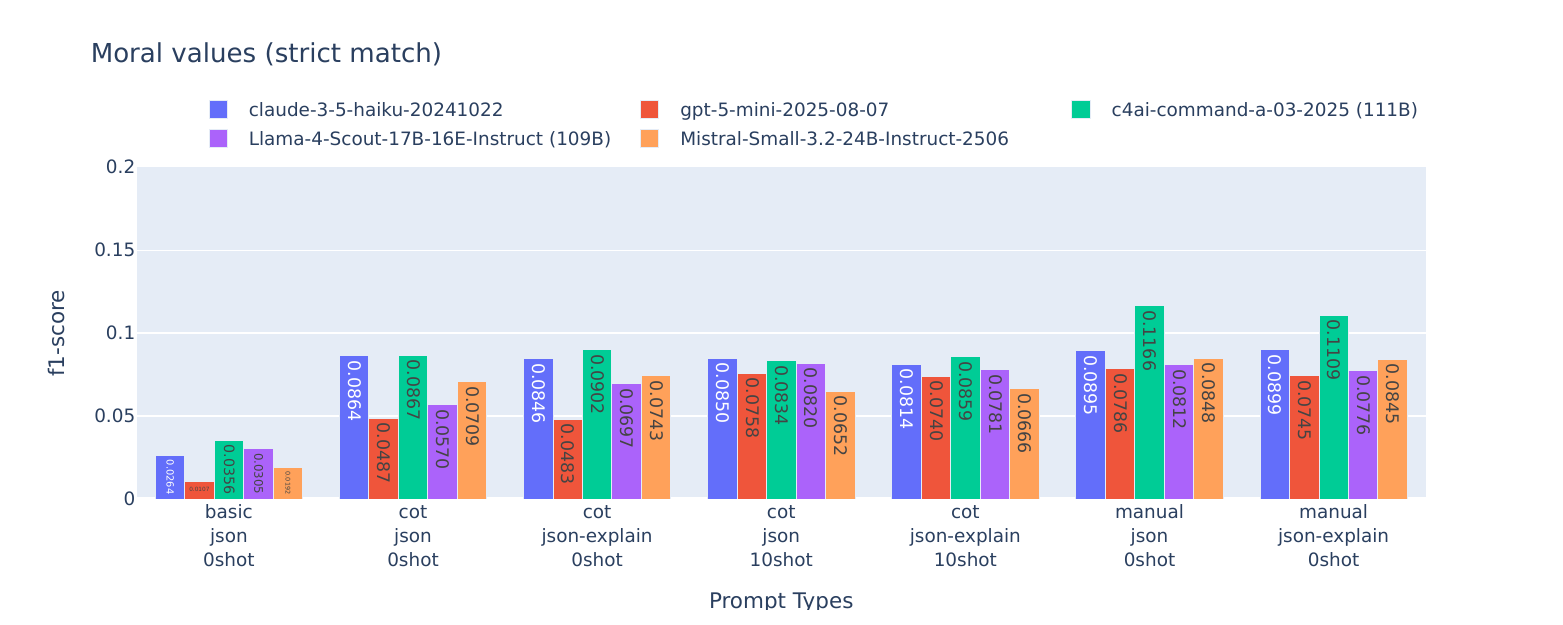}
    \caption{Moral Value evaluation with MFT category labels in the strict match criteria. Note that the multi-label annotation is allowed for a single moral value span.}
    \label{fig:eval1}
\end{figure*}

\begin{figure*}
    \centering
    \includegraphics[width=0.8\linewidth]{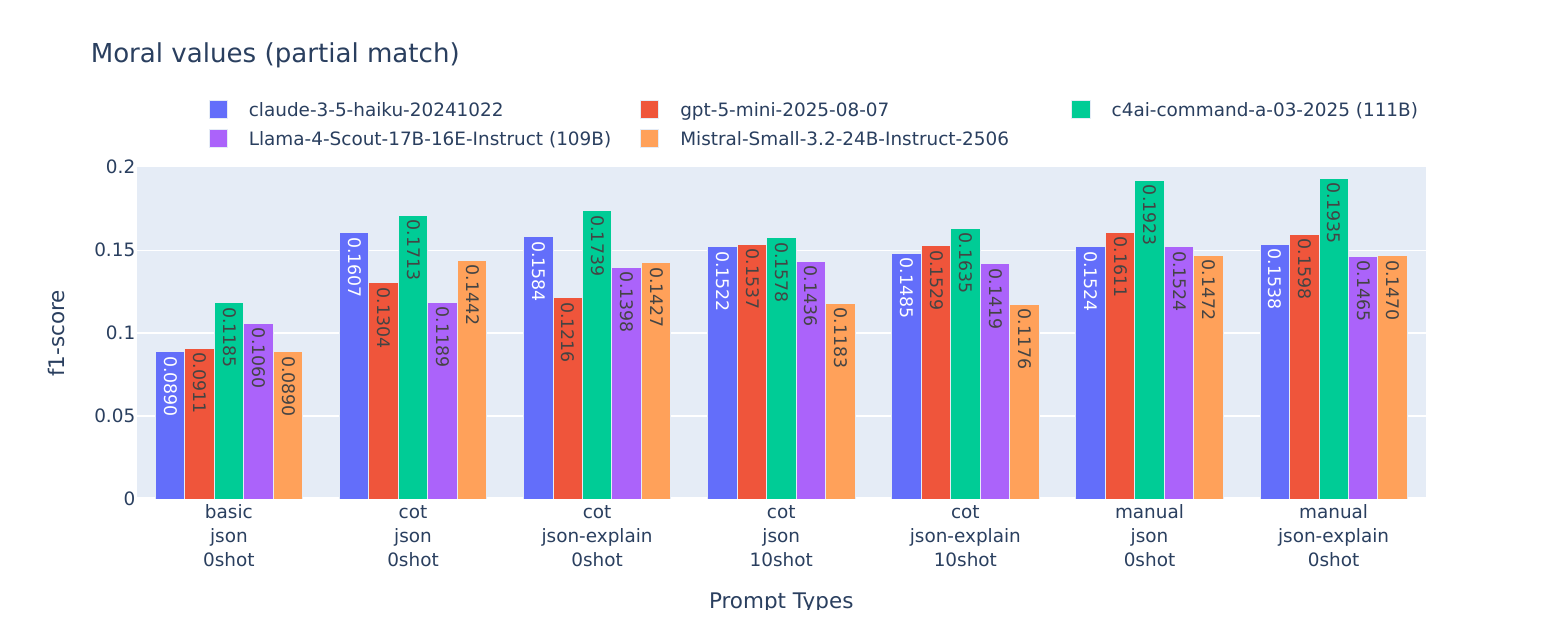}
    \caption{Moral Value evaluation with MFT category labels in the partial match criteria. Again, multi-label annotation is allowed for a single moral value span.}
    \label{fig:eval2}
\end{figure*}

\begin{figure*}
    \centering
    \includegraphics[width=0.8\linewidth]{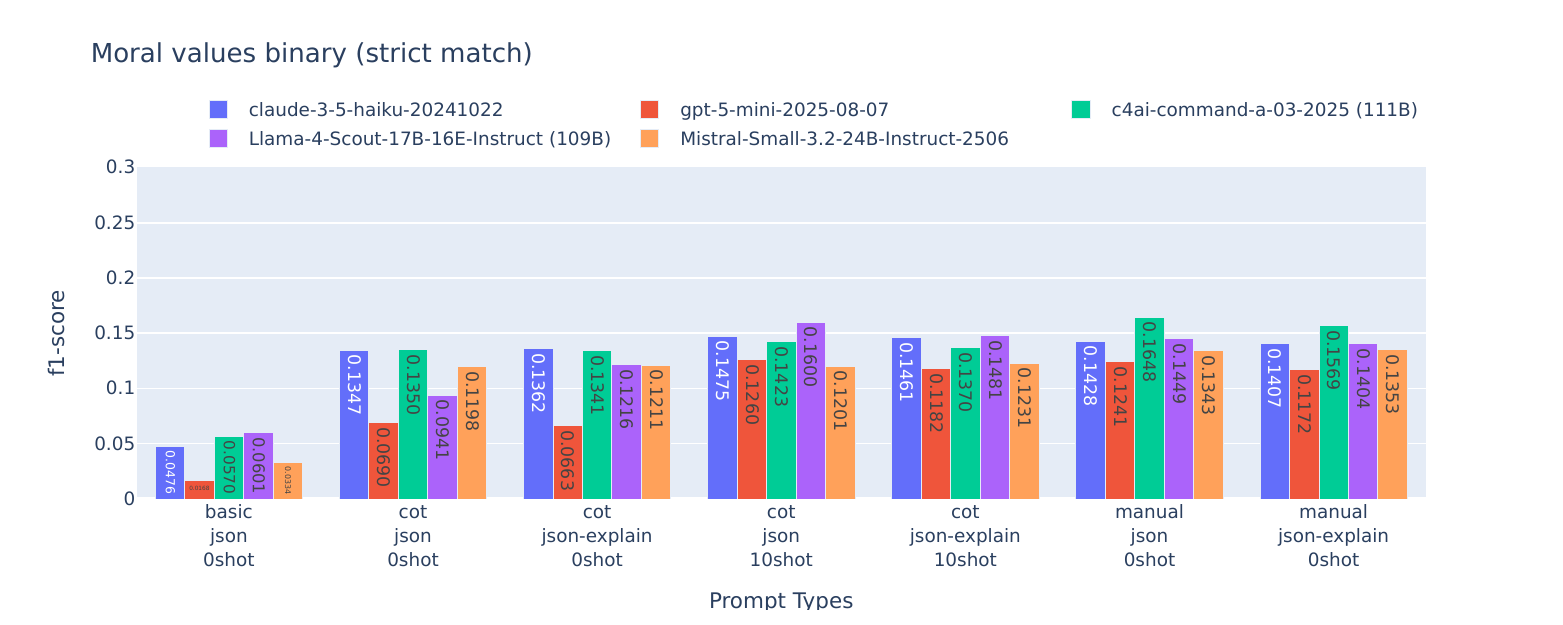}
    \caption{Moral Value evaluation with virtue/vice labels only in the strict match criteria. MFT category difference between gold and prediction is not taken into account. Note that we resolved the multi-label annotation onto multi-class (two-classes virtue or vice) per moral value.}
    \label{fig:eval3}
\end{figure*}

\begin{figure*}
    \centering
    \includegraphics[width=0.8\linewidth]{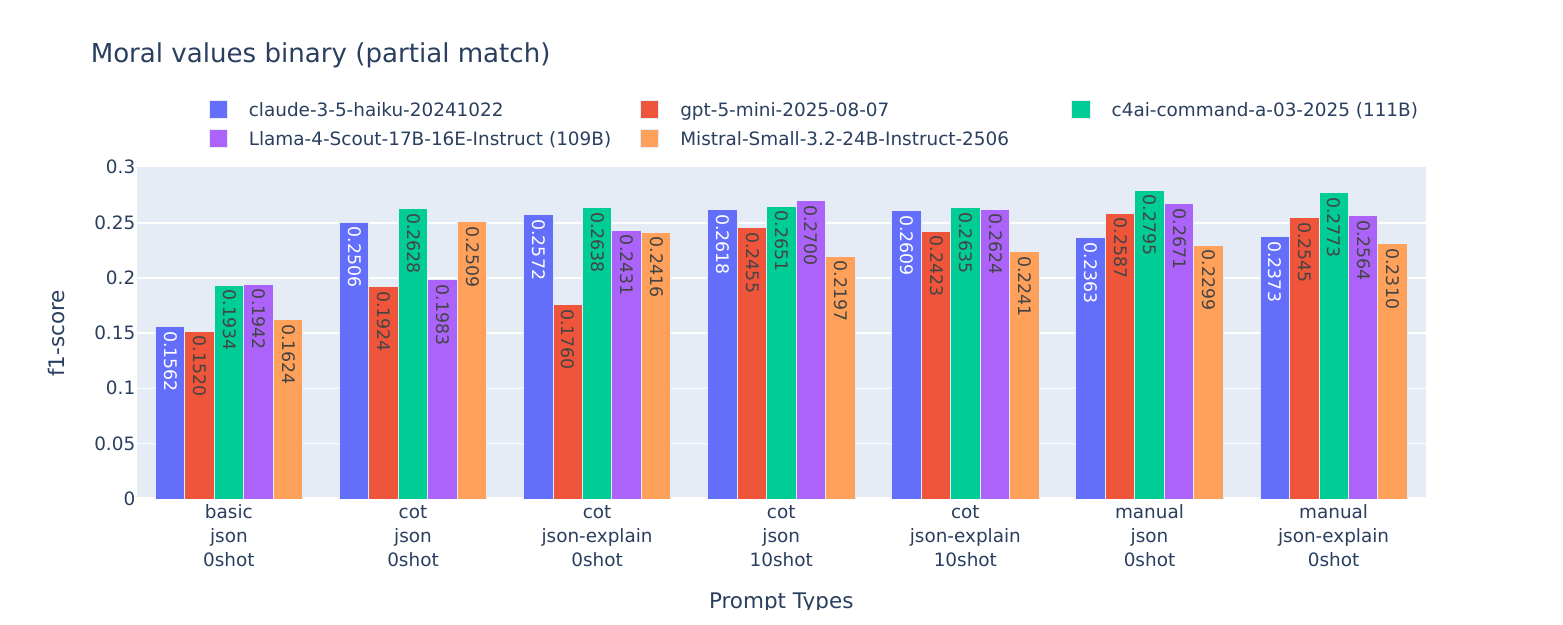}
    \caption{Moral Value evaluation with virtue/vice labels only in the partial match criteria.}
    \label{fig:eval4}
\end{figure*}

Fig.~\ref{fig:eval1}, \ref{fig:eval2}, \ref{fig:eval3}, and \ref{fig:eval4} show the detailed results for automatic moral value evaluation with the SemEval-2013 NER-style setup \cite{segura-bedmar-etal-2013-semeval} using strict and partial matching (see \S \ref{sec:eval-moral-components}).

\subsubsection{Protagonist Evaluation}
\label{appendix-pr}

\begin{figure*}
    \centering
    \includegraphics[width=0.8\linewidth]{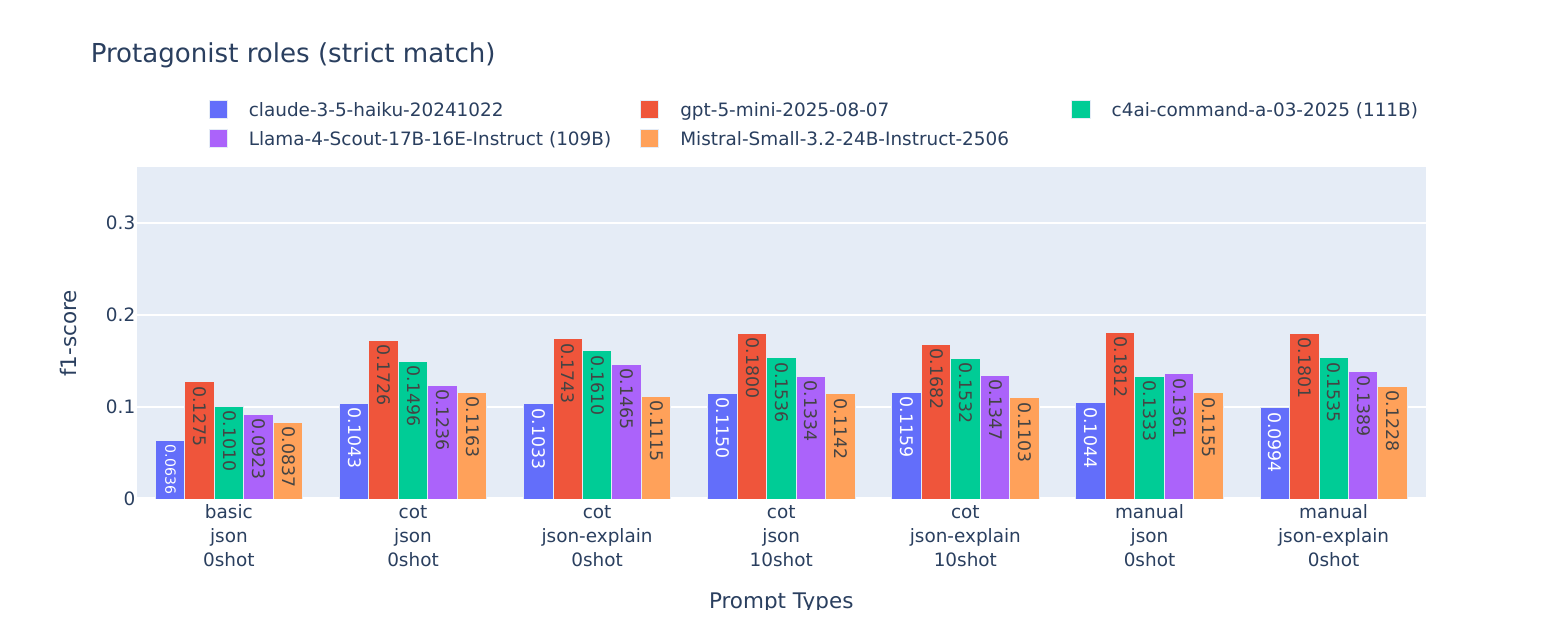}
    \caption{Protagonist evaluation with role labels in the strict match criteria. Multi-label annotation is allowed for a single protagonist span.}
    \label{fig:eval5}
\end{figure*}

\begin{figure*}
    \centering
    \includegraphics[width=0.8\linewidth]{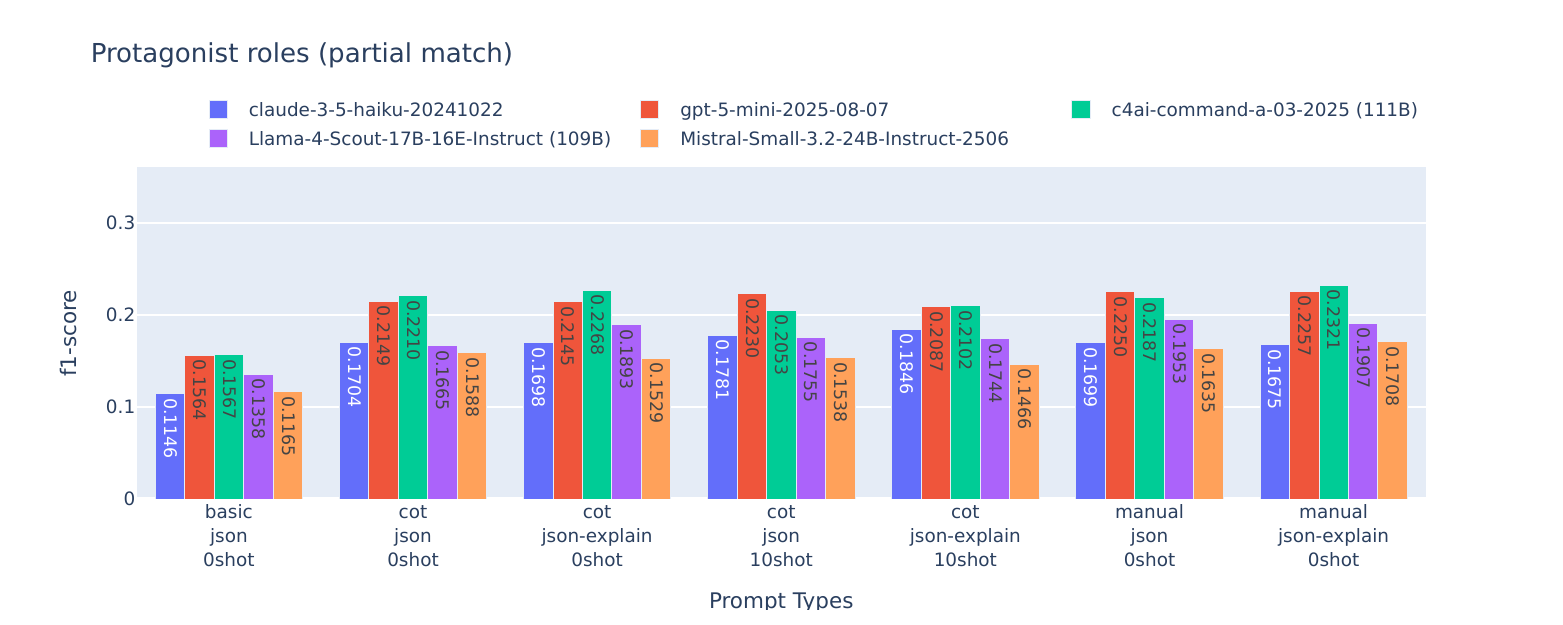}
    \caption{Protagonist evaluation with role labels in the partial match criteria. Especially in the partial match, we relaxed the criteria further for noun phrases, so that the start position difference (whether a preceding determinat is present in the span or not) does not affect much.}
    \label{fig:eval6}
\end{figure*}

\begin{figure*}
    \centering
    \includegraphics[width=0.8\linewidth]{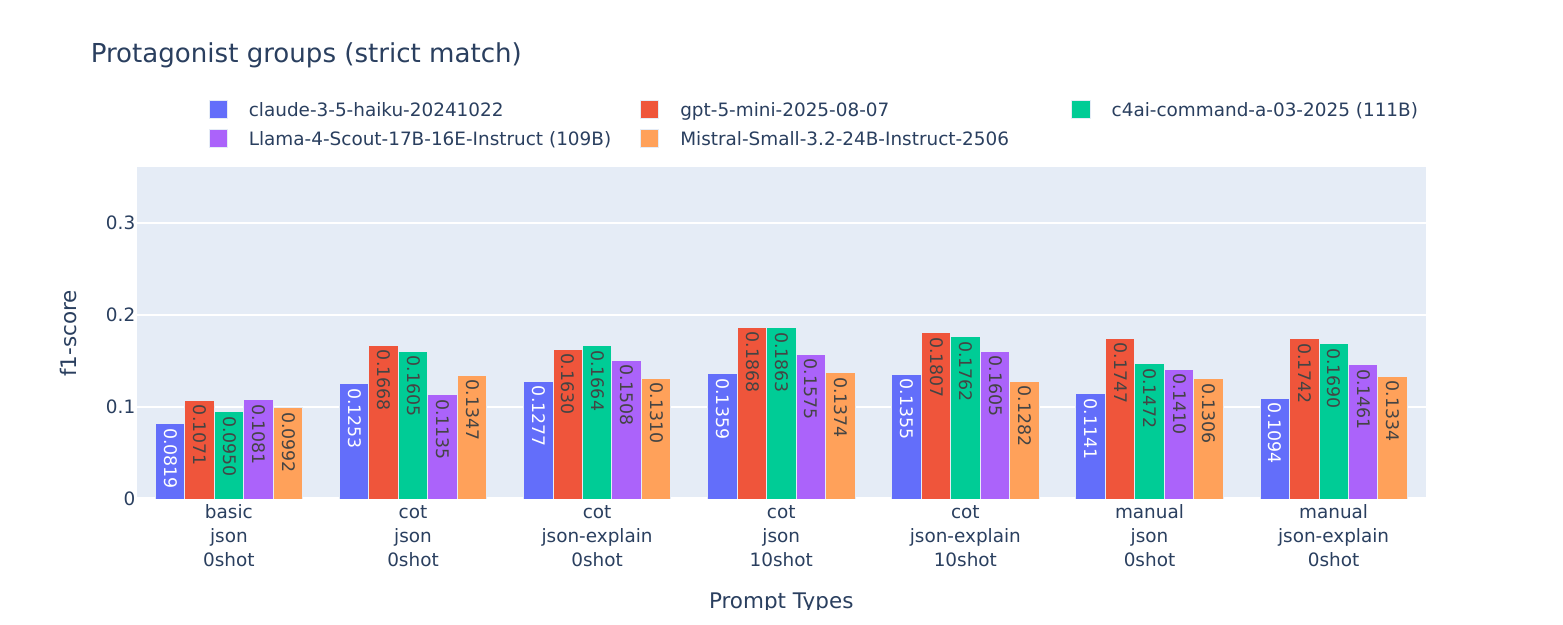}
    \caption{Protagonist evaluation with group labels in the strict match criteria. Note that the group labels are mutually exclusive, which means the original multi-class evaluation is applied without adjustment for multi-label setup.}
    \label{fig:eval7}
\end{figure*}

\begin{figure*}
    \centering
    \includegraphics[width=0.8\linewidth]{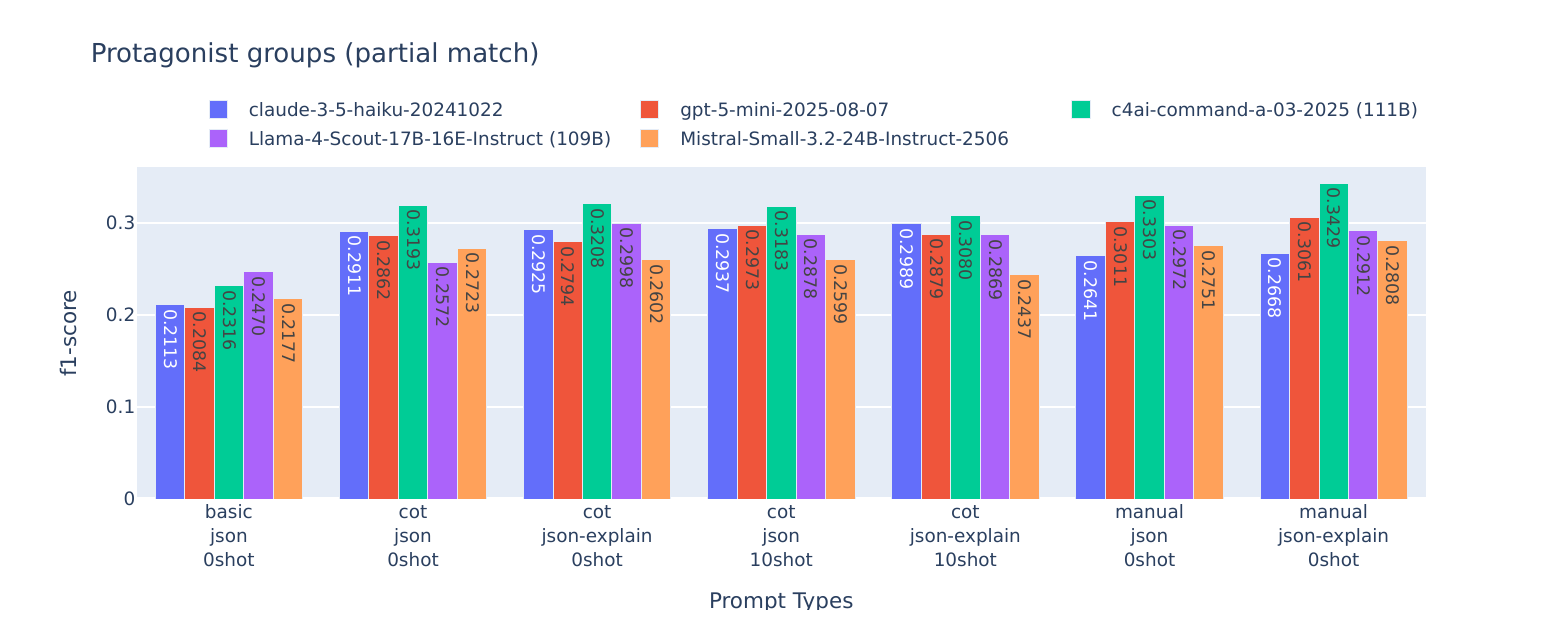}
    \caption{Protagonist evaluation with group labels in the partial match criteria. We applied the relaxed rule for noun phrases described above.}
    \label{fig:eval8}
\end{figure*}

\begin{figure*}
    \centering
    \includegraphics[width=0.8\linewidth]{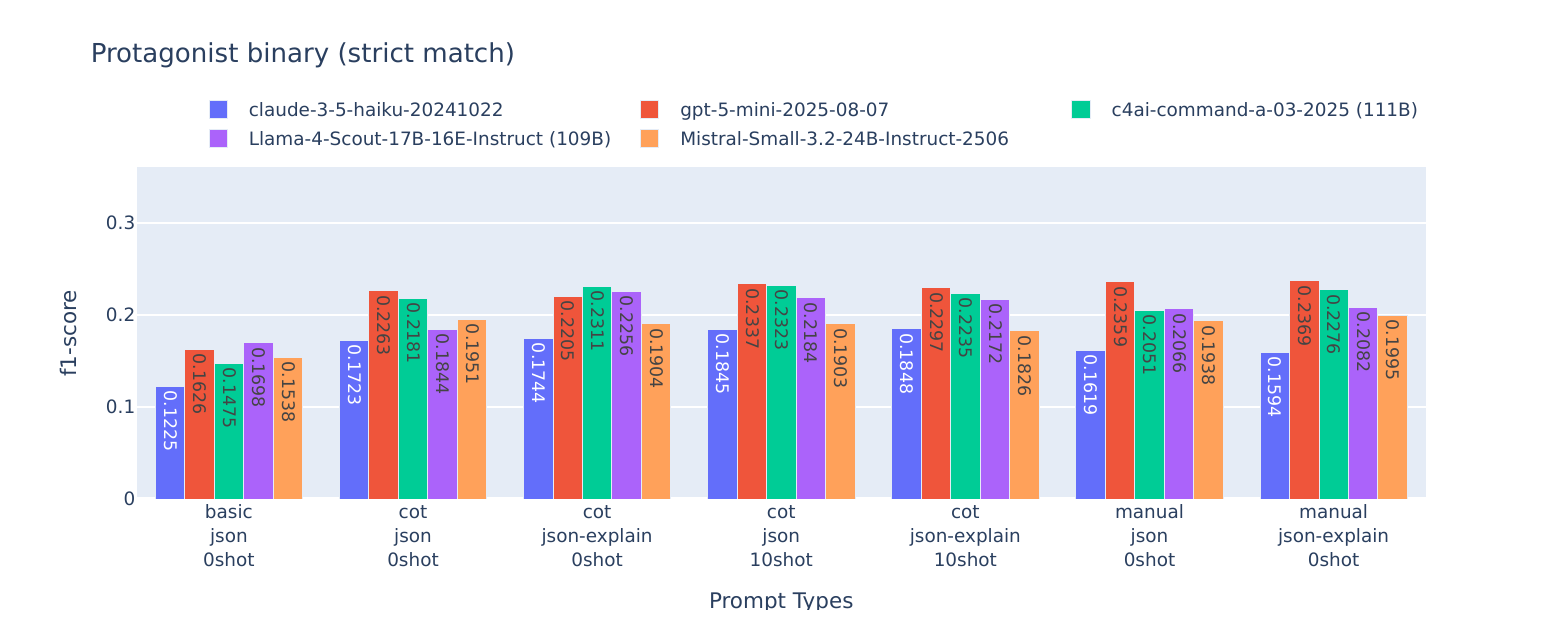}
    \caption{Protagonist evaluation without labels in the strict match criteria. We focus on the span start and end positions only, no label differences (neither role labels nor group labels) are taken into account.}
    \label{fig:eval9}
\end{figure*}

\begin{figure*}
    \centering
    \includegraphics[width=0.8\linewidth]{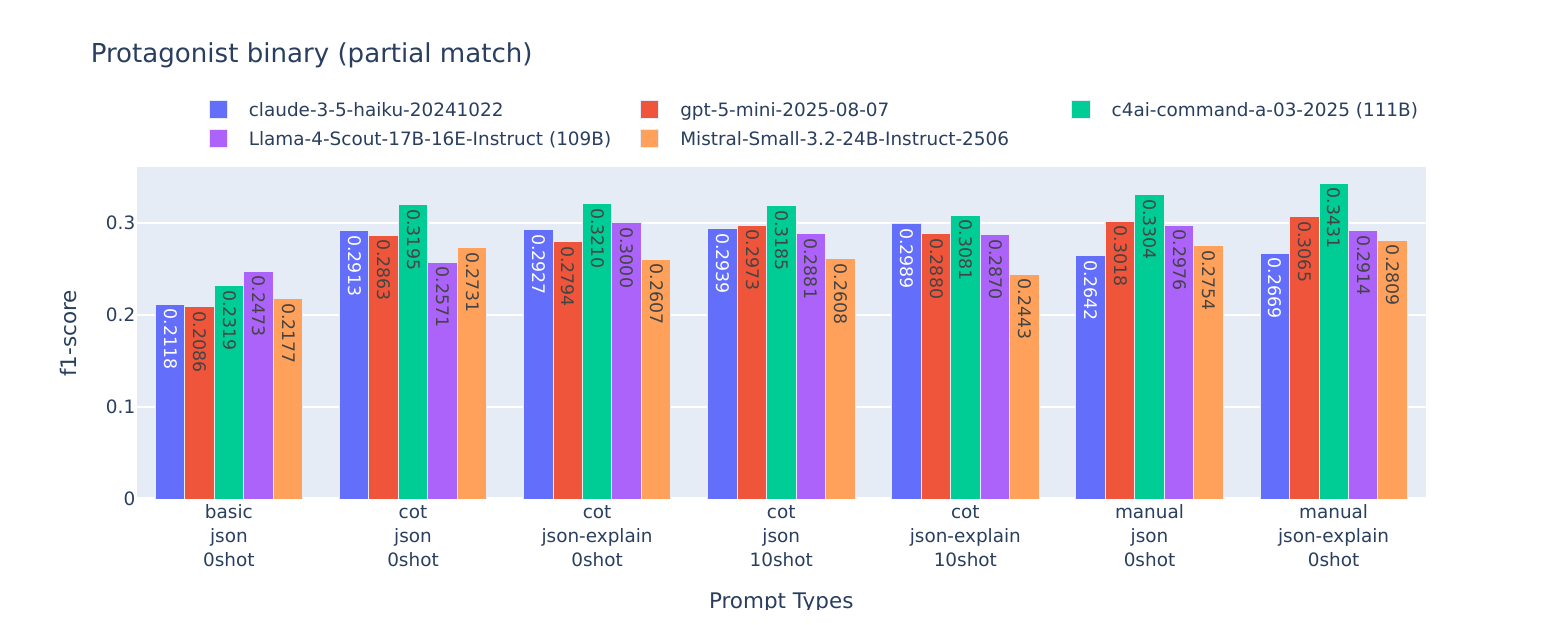}
    \caption{Protagonist evaluation without labels in the partial match criteria. Again, no label differences (neither role labels nor group labels) are taken into account.}
    \label{fig:eval10}
\end{figure*}

Fig. \ref{fig:eval5}, \ref{fig:eval6}, \ref{fig:eval7}, \ref{fig:eval8}, \ref{fig:eval9}, and \ref{fig:eval10} show the detailed results for automatic protagonist evaluation, again with the SemEval-2013 NER-style setup using strict and partial matching (see \S \ref{sec:eval-moral-components}).

\subsubsection{Demand Evaluation}

\begin{figure}
    \centering
    \includegraphics[width=1\linewidth]{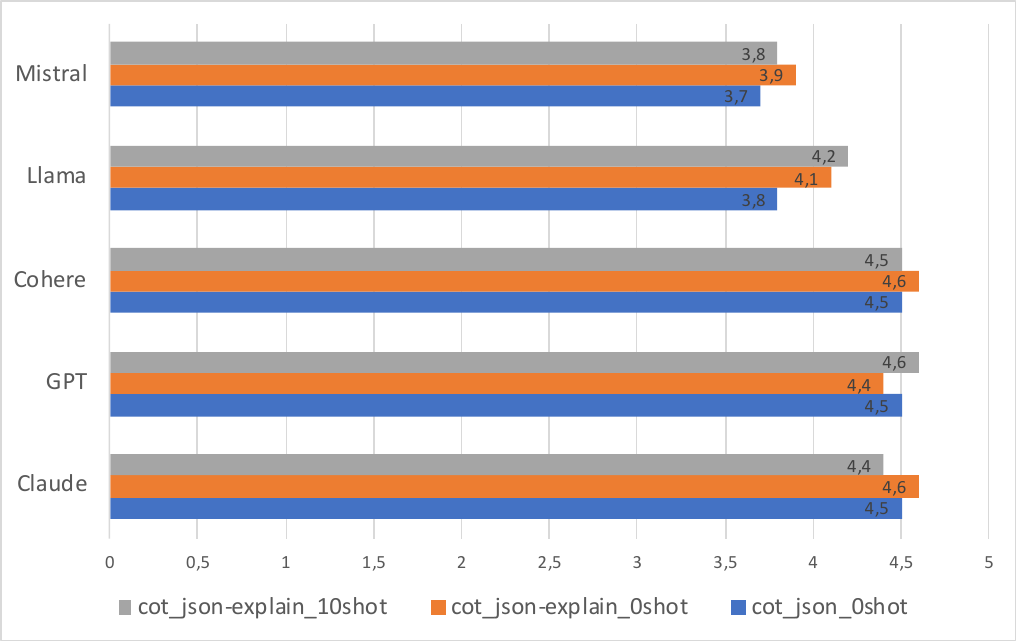}
    \caption{Manual evaluation of demands on a five-point Likert scale for semantic correctness, i.e., how accurately the demand conveyed the intended moral claim.}
    \label{fig:app-demand}
\end{figure}

Fig. \ref{fig:app-demand} shows the results of our manual evaluation of generated and extracted moral demands on Test-150 on a five-point Likert scale for semantic correctness (see \S \ref{sec:eval-moral-components}).

\subsection{Data Analysis}
\label{sec:appendix-ana}

\subsubsection{Agreement Scores for Test-150}

\begin{figure*}
    \centering
    \includegraphics[width=0.8\linewidth]{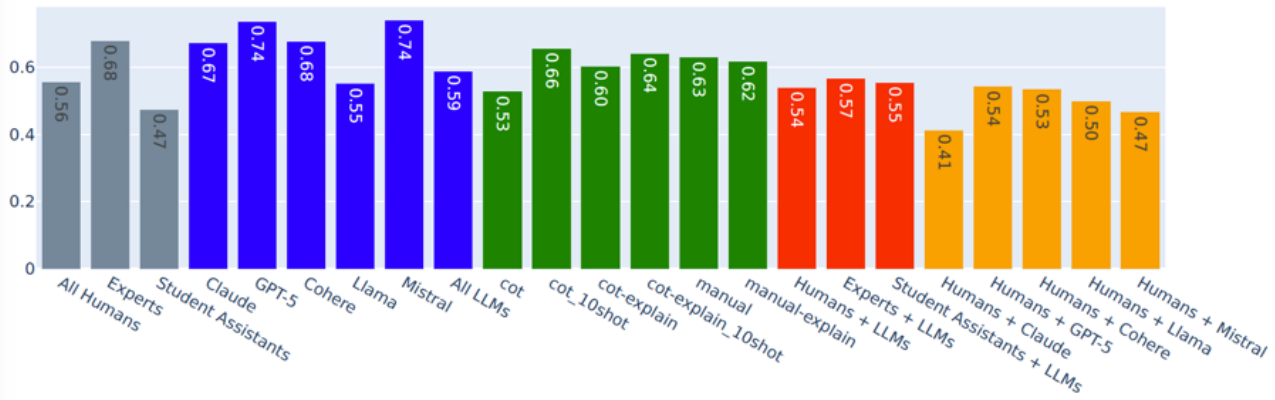}
    \caption{Fleiss' Kappa scores for different comparisons of agreement between and within human annotators and models.}
    \label{fig:kappa}
\end{figure*}

In \S \ref{sec:eval-agr} we compare human-human, model-model, and human-model agreement using PABAK (as displayed in the main paper) and Fleiss' Kappa as displayed in  Fig. \ref{fig:kappa}.

\subsubsection{Agreement between Models and Humans}
\label{appendix:agreement}

Table \ref{tab:agreement} displays the results of our agreement evaluation in \S \ref{sec:agreement}, focusing on the most clearly classified cases.

\begin{table}[]
    \centering
    \small
    \begin{tabular}{c|c|c|c}
    \toprule
Agreement & Models & Humans & M-H\\
\hline
Moralizations $\geq80$\% & 41.3 &22.0 & 20.7\\
Moralizations 100\% & 18.7 & 12.7  & 8.7\\
NVR $\geq80$\%  & 24.0  & 57.3 & 24.0 \\
NVR 100\% & 8 & 48 & 8\\
\bottomrule
    \end{tabular}
    \caption{Agreement and Disagreement among and between Models and Humans, in percentage.}
    \label{tab:agreement}
\end{table}

\subsubsection{Linguistic Indicators of Moralization}

To examine the hypothesis that specific textual features serve as key determinants of both model and human decisions, in \S \ref{sec:eval-ind} we conducted an analysis focusing on a selected set of linguistically motivated indicators. Results (in percentage) are displayed in Table~\ref{tab:indi}.

\begin{table*}
\small
\centering
\begin{tabular}{l|l|l|l|l|l}
\toprule
                     & \textbf{Moral Phrases} & \textbf{Explicit Demand} & \textbf{Modal Verb} & \textbf{Subjunctive} & \textbf{Context}  \\
                     \hline
Moralizations (n=31) & 28                            & 25                       & 22                         & 7                    & 0                         \\
Percentage           & 90.3                          & 8.6                      & 71                         & 22.6                 & 0                         \\
NVR (n=36)           & 9                             & 0                        & 8                          & 3                    & 11                        \\
Percent              & 25                            & 0                        & 22.2                       & 8.3                  & 30.6                   \\
\bottomrule
\end{tabular}
\caption{Textual features of moralizations and NVIs, numbers in percentage.}
\label{tab:indi}
\end{table*}

\subsubsection{Details on the Deviation Analysis}
\label{sec:details-deviation}

In \S \ref{sec:eval-var} we investigated possible sources of divergence between model and human decisions. In contrast to the previous analyses -- which compared patterns of human and machine agreement -- this analysis required a reference point. Therefore, model predictions were compared with the majority vote of the human annotators for Test-150 to identify typical deviations. It is important to emphasize that, in the context of inherently subjective annotation tasks such as moralization classification, deviations from the reference are not regarded as errors but as variance \cite{falk2025mining, chochlakis2024larger}. For ease of general comprehensibility, however, we refer to these cases as false positives (FPs) -- instances labeled by humans as NVIs but predicted as moralizations by a model, and false negatives (FNs) -- instances labeled by humans as moralizations but predicted as NVIs.

\textbf{Evaluation Setup.} In a bottom-up process, potential causes for divergences between model and human decisions were first explored on the entire test set (excluding Test-150) and then grouped into categories for FPs and FNs separately. Based on these observations, an annotation manual was developed, including detailed category definitions, examples, and annotation guidelines. Two research assistants with a background in linguistics conducted parallel manual annotations of all false positives and false negatives (across all five models and seven configurations) within Test-150. Annotations were performed blind, meaning the annotators were not informed about which model or configuration produced which prediction.\footnote{Annotators were, in fact, not told that the data originated from model predictions at all.}  We measured inter-annotator agreement (IAA) using Cohen's Kappa and achieved an agreement of 0.72\% for the false-positive categories and 0.69\% for the false-negative categories. Remaining disagreements were resolved by an expert annotator (one of the authors).

\textbf{Results for False Positives.}
For model deviations classified as false positives -- that is, instances labeled by humans as NVIs but predicted by a model as moralizations -- three main categories emerged:

\begin{enumerate}
    \item \textbf{Use of moral vocabulary in neutral contexts}: Many models label passages as moralizing whenever words such as duty, moral, responsibility, right, or wrong appear, even when used descriptively or within quotations. In our dataset, this often occurs in historical reporting, where frequent negative moral terms such as \textit{horrible epochs, time of plague, Thirty Years' War} appear, without constituting moralization as a discursive strategy in our sense. Neutral uses of moral vocabulary also appear in passages that weigh different moral perspectives without articulating a demand or persuasive stance, which by our definition likewise do not qualify as moralizing.
    \item \textbf{Lack of context}: In some cases, the surrounding context is missing, making it impossible to determine whether moralization is present.
    \item \textbf{Borderline cases}: Since moralization detection is a subjective task, certain instances can reasonably be interpreted either way. These ambiguous cases are those where moralization could plausibly be argued even though the majority human label is NVI.
\end{enumerate}

\begin{figure}[t]
\begin{centering}
\includegraphics[width=0.48
\textwidth]{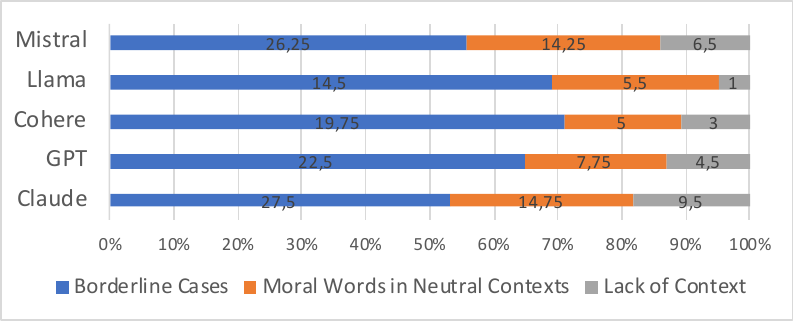}
\caption{Error causes for false positives, mean values (in total) per model across all prompt configurations.} 
\label{fig:fp}
\end{centering}
\end{figure}

Across all five models and seven configurations, Test-150 contains 729 false positives.\footnote{Accordingly, these findings are based on a relatively small dataset, and the scalability of the results should be validated in future work.} The distribution of categories (neutral / context / borderline) is visualized in Fig. \ref{fig:fp}. 

\textbf{Findings}. Models such as LLaMA, Cohere, and GPT -- which also achieved the best performance in the binary evaluation (see \S \ref{sec:data}) -- exhibit the highest proportion of borderline cases, suggesting that these systems may be more nuanced than raw metrics indicate. By contrast, Mistral and Claude, which performed weakest overall, struggle most with identifying neutral texts as non-moralizing, consistent with their lower precision scores. Missing context affects all models to a similar extent (except for LLaMA, which shows slightly greater robustness). This context problem, already discussed above as a potential signal for NVIs, here reappears as a factor influencing moralization decisions as well. Missing context thus represents a general limitation for moralization analysis and should be addressed in future work through targeted dataset expansion or context reconstruction.
When mapping these categories (neutral, context, borderline) back to the 12 cases where models and humans strongly diverge ($\geq80$\% of models vs. $\geq80$\% of humans giving opposing labels), 10 out of 12 are classified as borderline. This indicates that the cases in which humans and models diverge most are precisely those that are highly open to interpretation -- a plausible and consistent finding that warrants further validation with additional data.

\begin{figure}[t]
\begin{centering}
\includegraphics[width=0.48
\textwidth]{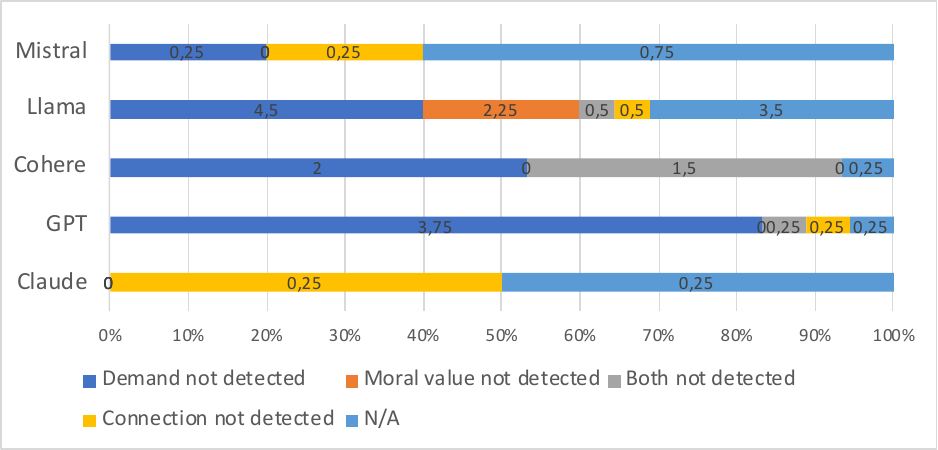}
\caption{Error causes for false negatives, mean values (in total) per model across all prompt configurations.} 
\label{fig:fn}
\end{centering}
\end{figure}

\textbf{Results for False Negatives.}
The second part of the analysis concerns false negatives, i.e., instances identified by human annotators as moralizations but not recognized as such by the models. The annotation parameters (deviation categories) were derived from the main defining features of moralizations (see \S \ref{sec:introduction}). The underlying hypothesis is that the models were prompted to first identify the key components of a moralization (moral values, demands, and their interrelation) and then make a classification decision based on this. We therefore assume that divergences from human judgments occur when at least one of these components is not recognized by the model. The categories used are (1) unrecognized moral word(s) or phrase(s); (2) unrecognized demand; (3) both unrecognized moral values and unrecognized demand; and (4) unrecognized connection between moral values and demand. The last category applies when a model correctly identifies both a moral value and a demand but still labels the instance as a NVI.\footnote{Although in rare cases moral values and demands may co-occur without an argumentative connection, we assume -- in line with the human majority label and for analytical simplicity -- that this is not the case in the present data.}

\textbf{Findings.} The analysis of false negatives (n = 85)\footnote{Given the limited data and low number of false negatives for some models/configurations, these findings should be interpreted with caution.} shows that the detection of demands poses a particular challenge even for the best-performing models (Cohere, LLaMA, GPT). A close inspection of the affected instances and their gold annotations reveals that 43 of 51 unrecognized demands are implicit, supporting our hypothesis that implicit moral demands are generally more difficult for language models to detect than explicit ones. Many of these cases involve descriptions of a negative (and thus implicitly undesirable) state of affairs without an explicit call for change.

Failure to detect moral vocabulary was a major source of divergence only for LLaMA and Cohere.\footnote{Cohere failed to detect both moral vocabulary and demands in 6 of its 15 false negatives.} Manual inspection revealed that this typically occurs when identifying moral terms requires world knowledge or sociocultural context. 

Failure to detect the connection between moral value and demand emerged as a key source of error particularly for LLaMA, Mistral, and Claude.\footnote{Given the small number of affected instances, caution is warranted when interpreting these results, especially for Claude and Mistral.} Apart from the 14 LLaMA cases falling into this category, only eight additional instances across models met this criterion -- a finding that supports the assumption that co-occurrence of moral vocabulary and demands within a short span usually indicates moralization. Example 20 illustrates a case in which models (Claude, Mistral, LLaMA) correctly detect moral components but fail to recognize that they serve as argumentative support for a demand.

Additional, more subtle sources of error emerged from manual inspection: Figurative language or irony can obscure moral content, leading models to miss moralizations; complex argument structures, particularly when the link between moral value and demand spans multiple sentences, pose difficulties; and negated demands, where a text argues that a certain moral action is not required (e.g., ``no need to make amends''), are often overlooked, even though such negations are constitutive of moralization by our definition.

\subsection{Examples}
\label{sec:appendix-exa}

For reasons of space, only a limited number of examples could be included in the paper itself. However, throughout the chapters, references were made to additional examples, which are listed below, organized according to the respective chapters. All translations by the authors of this paper.

\subsubsection{Illustrating Examples for \S \ref{sec:stats}}

\begin{small}\textbf{Example 1 (Wikipedia Discussions) – Moral demands
for neutral description in Wikipedia articles}: \textit{
In my view, a neutral description of reality
always means describing what was formative for
both the past and the present in a balanced way.}

\textbf{Example 2 (Wikipedia Discussions) – Moral
demands for neutral description in Wikipedia
articles:} \textit{I don’t think it’s the task of an encyclopedia
to express such judgments. If you criticize
that something isn’t written neutrally, then point out
the passages and explain what exactly isn’t neutral.
Just because you feel the article sounds too positive
doesn’t mean it is. . . maybe your own attitude
toward the topic isn’t neutral enough.}

\textbf{Example 3 (Letters to the editors) – Moralization
where the actual addressee is a diffuse
public audience:} \textit{Beckmann’s angry speech is
justified. It is outrageous how arrogantly our politicians
ignore the needs of our children. From a
political perspective, children do not pay off—but
for society they certainly do. Yet recognizing that
requires a certain degree of foresight.}

\textbf{Example 4 (Parliamentary debates) -- Moralization that serves to legitimize and morally reinforce desirable actions:} \textit{Anyone who wants to fight right-wing extremism must fundamentally revise and abolish xenophobic provisions and laws. When people from Vietnam who have lived here for more than ten years, or people from crisis regions like Sri Lanka, are deported, this must be addressed too—just like the ghettoization of refugees. This, in my view, plays directly into the hands of the arsonists, because inhibitions are being lowered.}

\textbf{Example 5 (Parliamentary debates) -- Moralization that emphasize positive social outcomes of moral action:} \textit{That is why it is not enough to simply stand up and say: We reject all forms of violence with disgust and indignation. We must not only show that we do not tolerate racism and violence, but also convey the democratic values by which we want to convince our children and the youth in both East and West. Therefore, I believe that under no circumstances should we restrict civil rights, such as the right to assembly. That would amount to a capitulation of the rule of law to right-wing extremists, ladies and gentlemen.}

\textbf{Example 6 (Parliamentary debates) -- Moralization where the beneficiary stays implicit:} \textit{
We need stricter laws to prevent racially motivated violence.}

\textbf{Example 7 (Letters to the editors) -- Moralization where generic expressions are used to evoke shared societal concerns:} \textit{
I would like to arrive at a definition that grants all people equal humanitarian protection—regardless of whether they are persecuted by the state or by non-state actors.}

\subsubsection{Illustrating Examples for \S \ref{sec:agreement}}

\textbf{Example 8 (Letters to the editor) -- Moralization with total agreement between all models and humans:} \textit{
Dear people, if you want to make a difference, then you should be consistent and vote out the responsible, established parties, boycott the products of the polluters and warmongers, and switch to fair trade. Immediately and rigorously align your own life ecologically and economically, and fight for a just world. Only then will anything improve.}

\textbf{Example 9 (Nonfiction books) -- NVI with total agreement between all models and humans:} \textit{ 
On the eve of the First World War, faith in progress was accompanied by fear of the abyss. Nietzsche had become the herald of the apocalyptic spirit of the age, who in Ecce Homo spoke of wars ``such as have never yet existed on earth.'' It was into such a war — in which ``the nineteenth century [...] was shot to pieces'' (AH1/SW9, 133) and the ideal of eternal peace destroyed — that Ernst Juenger was to go.}

\textbf{Example 10 (Letters to the editor) -- 33 out of 35 models predict a moralization, 4 out of 5 humans annotate a NFR:} \textit{
The penal order procedure may be suitable for tax evaders. For those who make up the bulk of the accused, however, the opportunity for an oral hearing is necessary. With the proliferation of written procedures, the judiciary has become alienated from the citizens. Yet without a simplification of legal language, even the strengthening of oral hearings will not achieve much. The mechanically recited indictment has degenerated into an undignified ritual. }

\subsubsection{Illustrating Examples for \S \ref{sec:eval-var}}

\textbf{Example 11 (Nonfiction Books) -- Use of moral vocabulary in neutral contexts:} \textit{
Beneath the smoking ruins lay around seventy million dead. In breathtaking recklessness, the politicians had unleashed the dogs of war and triggered a frenzy of self-destruction. History has known terrible epochs such as the time of the plague or the Thirty Years' War, but never before had there been massacres of such magnitude as in the Thirty Years' War of 1914 to 1945 (setting aside the brief pause in between).}

\textbf{Example 12 (Letters to the editor) -- Missing Context:} \textit{
The explicit goal of the award ceremony was, not least, to bring the forgotten fate of this endangered people before the world public and the conscience of humanity. }

\textbf{Example 13 (Parliamentary Debates) -- Borderline Cases:} \textit{
Empirical evidence is sometimes more valuable than the occasional utopian idea, no matter how often it has been written down. That is why we are well advised to build on what Ms. Kotting-Uhl has introduced here. It is forestry from which the concept of sustainability originally stems — developed not so much out of a love of trees and forests, but from a purely pragmatic pursuit of profit. This again shows that ecology and economy can go very well together.}

\end{small}

%% file: custom.bib
@inproceedings{becker-etal-2024-detecting,
    title = "Detecting Impact Relevant Sections in Scientific Research",
    author = "Becker, Maria  and
      Han, Kanyao  and
      Lee, Haejin  and
      Werthmann, Antonina  and
      Rezapour, Rezvaneh  and
      Diesner, Jana  and
      Witt, Andreas",
    editor = "Calzolari, Nicoletta  and
      Kan, Min-Yen  and
      Hoste, Veronique  and
      Lenci, Alessandro  and
      Sakti, Sakriani  and
      Xue, Nianwen",
    booktitle = "Proceedings of the 2024 Joint International Conference on Computational Linguistics, Language Resources and Evaluation (LREC-COLING 2024)",
    month = may,
    year = "2024",
    address = "Torino, Italia",
    publisher = "ELRA and ICCL",
    url = "https://aclanthology.org/2024.lrec-main.424/",
    pages = "4744--4749"
}

@inproceedings{chochlakis2024larger,
  title = {Larger Language Models Don't Care How You Think: Why Chain-of-Thought Prompting Fails in Subjective Tasks},
  author = {Georgios Chochlakis and Niyantha Maruthu Pandiyan and Kristina Lerman and Shrikanth Narayanan},
  booktitle = {Proceedings of the IEEE International Conference on Acoustics, Speech and Signal Processing (ICASSP)},
  year = {2025},
  url = {https://arxiv.org/abs/2409.06173}
}

@inproceedings{alvarez2024moral,
  title = {Moral Disagreement over Serious Matters: Discovering the Knowledge Hidden in the Perspectives},
  author = {Anny D. Alvarez Nogales and Oscar Araque},
  booktitle = {Proceedings of the 3rd Workshop on Perspectivist Approaches to NLP (NLPerspectives) @ LREC-COLING},
  year = {2024},
  pages = {67--77},
  url = {https://aclanthology.org/2024.nlperspectives-1.7/}
}

@inproceedings{weber2024varierr,
  title = {VariErr NLI: Separating Annotation Error from Human Label Variation},
  author = {Leon Weber-Genzel and Siyao Peng and Marie-Catherine De Marneffe and Barbara Plank},
  booktitle = {Proceedings of the 62nd Annual Meeting of the Association for Computational Linguistics (Volume 1: Long Papers)},
  year = {2024},
  pages = {2256--2269},
  url = {https://aclanthology.org/2024.acl-long.123/}
}

@inproceedings{falk2025mining,
  title = {Mining the Uncertainty Patterns of Humans and Models in the Annotation of Moral Foundations and Human Values},
  author = {Neele Falk and Gabriella Lapesa},
  booktitle = {Proceedings of the 63rd Annual Meeting of the Association for Computational Linguistics (Volume 1: Long Papers)},
  year = {2025},
  pages = {22898--22921},
  url = {https://aclanthology.org/2025.acl-long.1116/}
}

@inproceedings{rehbein-etal-2025-moral,
    title = "Moral reckoning: How reliable are dictionary-based methods for examining morality in text?",
    author = "Rehbein, Ines  and
      Brauner, Lilly  and
      Ertz, Florian  and
      Reinig, Ines  and
      Ponzetto, Simone",
    editor = {H{\"a}m{\"a}l{\"a}inen, Mika  and
      {\"O}hman, Emily  and
      Bizzoni, Yuri  and
      Miyagawa, So  and
      Alnajjar, Khalid},
    booktitle = "Proceedings of the 5th International Conference on Natural Language Processing for Digital Humanities",
    month = may,
    year = "2025",
    address = "Albuquerque, USA",
    publisher = "Association for Computational Linguistics",
    url = "https://aclanthology.org/2025.nlp4dh-1.20/",
    doi = "10.18653/v1/2025.nlp4dh-1.20",
    pages = "232--250",
    ISBN = "979-8-89176-234-3"
}

@misc{Landowska_Budzynska_Zhang_2024, title={Quantitative and qualitative analysis of moral foundations in argumentation - argumentation}, url={https://link.springer.com/article/10.1007/s10503-024-09636-x}, journal={SpringerLink}, publisher={Springer Netherlands}, author={Landowska, Alina and Budzynska, Katarzyna and Zhang, He}, year={2024}, month={Jul}}

@inproceedings{roy-etal-2022-towards,
    title = "Towards Few-Shot Identification of Morality Frames using In-Context Learning",
    author = "Roy, Shamik  and
      Nakshatri, Nishanth Sridhar  and
      Goldwasser, Dan",
    editor = "Bamman, David  and
      Hovy, Dirk  and
      Jurgens, David  and
      Keith, Katherine  and
      O'Connor, Brendan  and
      Volkova, Svitlana",
    booktitle = "Proceedings of the Fifth Workshop on Natural Language Processing and Computational Social Science (NLP+CSS)",
    month = nov,
    year = "2022",
    address = "Abu Dhabi, UAE",
    publisher = "Association for Computational Linguistics",
    url = "https://aclanthology.org/2022.nlpcss-1.20/",
    doi = "10.18653/v1/2022.nlpcss-1.20",
    pages = "183--196"
}

@inproceedings{zhang-etal-2024-moka,
    title = "{MOKA}: Moral Knowledge Augmentation for Moral Event Extraction",
    author = "Zhang, Xinliang Frederick  and
      Wu, Winston  and
      Beauchamp, Nick  and
      Wang, Lu",
    editor = "Duh, Kevin  and
      Gomez, Helena  and
      Bethard, Steven",
    booktitle = "Proceedings of the 2024 Conference of the North American Chapter of the Association for Computational Linguistics: Human Language Technologies (Volume 1: Long Papers)",
    month = jun,
    year = "2024",
    address = "Mexico City, Mexico",
    publisher = "Association for Computational Linguistics",
    url = "https://aclanthology.org/2024.naacl-long.252/",
    doi = "10.18653/v1/2024.naacl-long.252",
    pages = "4481--4502"
}

@inproceedings{lei-etal-2024-emona,
    title = "{EMONA}: Event-level Moral Opinions in News Articles",
    author = "Lei, Yuanyuan  and
      Miah, Md Messal Monem  and
      Qamar, Ayesha  and
      Reddy, Sai Ramana  and
      Tong, Jonathan  and
      Xu, Haotian  and
      Huang, Ruihong",
    editor = "Duh, Kevin  and
      Gomez, Helena  and
      Bethard, Steven",
    booktitle = "Proceedings of the 2024 Conference of the North American Chapter of the Association for Computational Linguistics: Human Language Technologies (Volume 1: Long Papers)",
    month = jun,
    year = "2024",
    address = "Mexico City, Mexico",
    publisher = "Association for Computational Linguistics",
    url = "https://aclanthology.org/2024.naacl-long.293/",
    doi = "10.18653/v1/2024.naacl-long.293",
    pages = "5239--5251"
}

@inproceedings{tubiblio106270,
            note = {Veranstaltungstitel: The 27th International Conference on Computational Linguistics (COLING 2018)},
        language = {en},
          author = {Jan-Christoph Klie and Michael Bugert and Beto Boullosa and Richard Eckart de Castilho and Iryna Gurevych},
           title = {The INCEpTION Platform: Machine-Assisted and Knowledge-Oriented Interactive Annotation},
            year = {2018},
           month = {Juni},
           pages = {5--9},
       eventdate = {20.08.2018-26.08.2018},
       publisher = {Association for Computational Linguistics},
        location = {Santa Fe, USA},
       booktitle = {Proceedings of the 27th International Conference on Computational Linguistics: System Demonstrations},
             url = {http://tubiblio.ulb.tu-darmstadt.de/106270/}
}

@article{Jiang2021DelphiTM,
  title={Delphi: Towards Machine Ethics and Norms},
  author={Liwei Jiang and Jena D. Hwang and Chandra Bhagavatula and Ronan Le Bras and Maxwell Forbes and Jon Borchardt and Jenny T Liang and Oren Etzioni and Maarten Sap and Yejin Choi},
  journal={ArXiv},
  year={2021},
  volume={abs/2110.07574},
  url={https://api.semanticscholar.org/CorpusID:238857096}
}

@article{Diakopoulos2014IdentifyingAA,
  title={Identifying and Analyzing Moral Evaluation Frames in Climate Change Blog Discourse},
  author={Nicholas A. Diakopoulos and Amy X. Zhang and Dag Elgesem and Andrew Salway},
  journal={Proceedings of the International AAAI Conference on Web and Social Media},
  year={2014},
  url={https://api.semanticscholar.org/CorpusID:935781}
}

@INPROCEEDINGS{10021123,
  author={Islam, Tunazzina and Goldwasser, Dan},
  booktitle={2022 IEEE International Conference on Big Data (Big Data)}, 
  title={Understanding COVID-19 Vaccine Campaign on Facebook using Minimal Supervision}, 
  year={2022},
  volume={},
  number={},
  pages={585-595},
  doi={10.1109/BigData55660.2022.10021123}}

@article{doi:10.1080/19331681.2013.826613,
author = {Morteza Dehghani, Kenji Sagae, Sonya Sachdeva and Jonathan Gratch},
title = {Analyzing Political Rhetoric in Conservative and Liberal Weblogs Related to the Construction of the “Ground Zero Mosque”},
journal = {Journal of Information Technology \& Politics},
volume = {11},
number = {1},
pages = {1--14},
year = {2014},
publisher = {Routledge},
doi = {10.1080/19331681.2013.826613},
URL = {https://doi.org/10.1080/19331681.2013.826613},
eprint = {https://doi.org/10.1080/19331681.2013.826613}
}

@inproceedings{reinig-etal-2024-survey,
    title = "A Survey on Modelling Morality for Text Analysis",
    author = "Reinig, Ines  and
      Becker, Maria  and
      Rehbein, Ines  and
      Ponzetto, Simone",
    editor = "Ku, Lun-Wei  and
      Martins, Andre  and
      Srikumar, Vivek",
    booktitle = "Findings of the Association for Computational Linguistics ACL 2024",
    month = aug,
    year = "2024",
    address = "Bangkok, Thailand and virtual meeting",
    publisher = "Association for Computational Linguistics",
    url = "https://aclanthology.org/2024.findings-acl.245",
    pages = "4136--4155",
}

@inproceedings{10.1145/2858036.2858423,
author = {Zhang, Amy X. and Counts, Scott},
title = {Gender and Ideology in the Spread of Anti-Abortion Policy},
year = {2016},
isbn = {9781450333627},
publisher = {Association for Computing Machinery},
address = {New York, NY, USA},
url = {https://doi.org/10.1145/2858036.2858423},
doi = {10.1145/2858036.2858423},
booktitle = {Proceedings of the 2016 CHI Conference on Human Factors in Computing Systems},
pages = {3378–3389},
numpages = {12},
location = {San Jose, California, USA},
series = {CHI '16}
}

@inproceedings{kiesel-etal-2023-semeval,
    title = "{S}em{E}val-2023 Task 4: {V}alue{E}val: Identification of Human Values Behind Arguments",
    author = "Kiesel, Johannes  and
      Alshomary, Milad  and
      Mirzakhmedova, Nailia  and
      Heinrich, Maximilian  and
      Handke, Nicolas  and
      Wachsmuth, Henning  and
      Stein, Benno",
    editor = {Ojha, Atul Kr.  and
      Do{\u{g}}ru{\"o}z, A. Seza  and
      Da San Martino, Giovanni  and
      Tayyar Madabushi, Harish  and
      Kumar, Ritesh  and
      Sartori, Elisa},
    booktitle = "Proceedings of the 17th International Workshop on Semantic Evaluation (SemEval-2023)",
    month = jul,
    year = "2023",
    address = "Toronto, Canada",
    publisher = "Association for Computational Linguistics",
    url = "https://aclanthology.org/2023.semeval-1.313",
    doi = "10.18653/v1/2023.semeval-1.313",
    pages = "2287--2303",
}

@article{haidt-etal-2009,
  author = {Jonathan Haidt and Jesse Graham and Conrad Joseph},
  year = 2009,
  title = {Above and below left–right: Ideological narratives and moral foundations},
  journal = {Psychological Inquiry}, 
  volume = 20,
  number = {2-3},
  pages = {110--119},
}

@incollection{graham-etal-2013,
title = {Chapter Two - Moral Foundations Theory: The Pragmatic Validity of Moral Pluralism},
editor = {Patricia Devine and Ashby Plant},
booktitle = {Advances in Experimental Social Psychology},
publisher = {Academic Press},
volume = {47},
pages = {55-130},
year = {2013},
issn = {0065-2601},
doi = {https://doi.org/10.1016/B978-0-12-407236-7.00002-4},
url = {https://www.sciencedirect.com/science/article/pii/B9780124072367000024},
author = {Jesse Graham and Jonathan Haidt and Sena Koleva and Matt Motyl and Ravi Iyer and Sean P. Wojcik and Peter H. Ditto}
}

@article{graham-etal-2009, 
    author = {Graham, Jesse and Jonathan Haidt and Brian A Nosek},
    year = 2009,
    title = {Liberals and conservatives rely on different sets of moral foundations},
    journal = {Journal of Personality and Social Psychology},
    volume = 96,
    number = 5,
    pages = {1029--1046},
    doi = {https://doi.org/10.1037/a0015141},
}

@article{schramowski2022large,
        author = {Patrick Schramowski and Cigdem Turan and Nico Andersen},
        title = {Large pre-trained language models contain human-like biases of what is right and wrong to do},
        journal = {Nature Machine Intelligence},
        volume = 4,
        pages = {258--268},
        year = {2022},
        doi = {https://doi.org/10.1038/s42256-022-00458-8},
}

@inproceedings{haemmerl-etal-2023-speaking,
    title = "Speaking Multiple Languages Affects the Moral Bias of Language Models",
    author = "Haemmerl, Katharina  and
      Deiseroth, Bjoern  and
      Schramowski, Patrick  and
      Libovick{\'y}, Jind{\v{r}}ich  and
      Rothkopf, Constantin  and
      Fraser, Alexander  and
      Kersting, Kristian",
    editor = "Rogers, Anna  and
      Boyd-Graber, Jordan  and
      Okazaki, Naoaki",
    booktitle = "Findings of the Association for Computational Linguistics: ACL 2023",
    month = jul,
    year = "2023",
    address = "Toronto, Canada",
    publisher = "Association for Computational Linguistics",
    url = "https://aclanthology.org/2023.findings-acl.134",
    doi = "10.18653/v1/2023.findings-acl.134",
    pages = "2137--2156",
}

@inproceedings{hendrycks2021aligning,
  title={Aligning AI With Shared Human Values},
  author={Hendrycks, Dan and Burns, Collin and Basart, Steven and Critch, Andrew Critch and Li, Jerry Li and Song, Dawn and Steinhardt, Jacob},
  booktitle={International Conference on Learning Representations},
  year={2021}
}

@article{mooijman2018moralization,
  title={Moralization in social networks and the emergence of violence during protests},
  author={Mooijman, Marlon and Hoover, Joe and Lin, Ying and Ji, Heng and Dehghani, Morteza},
  journal={Nature human behaviour},
  volume={2},
  number={6},
  pages={389--396},
  year={2018},
  publisher={Nature Publishing Group UK London}
}

@article{kampf2016political,
  title={Political Condemnations},
  author={Kampf, Zohar and Katriel, Tamar},
  journal={The Handbook of Communication in Cross-cultural Perspective},
  pages={312},
  year={2016},
  publisher={Taylor \& Francis}
}

@article{rhee2019and,
  title={The what, how, and why of moralization: A review of current definitions, methods, and evidence in moralization research},
  author={Rhee, Joshua J and Schein, Chelsea and Bastian, Brock},
  journal={Social and Personality Psychology Compass},
  volume={13},
  number={12},
  pages={e12511},
  year={2019},
  publisher={Wiley Online Library}
}

@misc{becker2023moral,
  title={The moral dimensions of health: pilot study},
  author={Becker, Maria and Ananth, Swetha and Kiemes, Carina},
  year={2023},
  publisher={Discourse Lab},
  doi={10.58079/no7q},
}

@book{belica2011semantische,
  title={Semantische N{\"a}he als {\"A}hnlichkeit von Kookkurrenzprofilen},
  author={Belica, Cyril},
  year={2011},
  publisher={Bozen University Press}
}

@article{felder2022diskurs,
  title={Diskurs korpuspragmatisch: Annotation, {K}ollaboration, {D}eutung am {B}eispiel von {P}raktiken des {M}oralisierens},
  author={Felder, Ekkehard and M{\"u}ller, Marcus},
  journal={Sprache in Politik und Gesellschaft: Perspektiven und Zug{\"a}nge},
  pages={241--262},
  year={2022},
  publisher={Walter de Gruyter GmbH \& Co KG}
}

@misc{ids2022dereko,
  author       = {{IDS}},
  title        = {Deutsches {R}eferenzkorpus / {A}rchiv der {K}orpora geschriebener {G}egenwartssprache 2022-{I} (Release vom 08.03.2022)},
  year         = {2022},
  address      = {Mannheim},
  publisher    = {Leibniz-Institut für Deutsche Sprache},
  note         = {PID: 00-04B6-B898-AD1A-8101-4},
  url          = {https://www1.ids-mannheim.de/kl/projekte/korpora}
}

@incollection{becker2025diskursgrammatik,
  author       = {Becker, Maria},
  title        = {Die {R}olle der {D}iskursgrammatik bei der {D}etektion und {A}nalyse sprachlicher {P}raktiken},
  booktitle    = {Diskursgrammatik},
  editor       = {Müller, M. and Reisigl, M. and Becker, M. and Bender, M. and Felder, E.},
  address      = {Berlin/Boston},
  publisher    = {De Gruyter},
  year         = {2025}
}

@misc{zhang2020bertscore,
      title={BERTScore: Evaluating Text Generation with BERT}, 
      author={Tianyi Zhang and Varsha Kishore and Felix Wu and Kilian Q. Weinberger and Yoav Artzi},
      year={2020},
      eprint={1904.09675},
      archivePrefix={arXiv},
      primaryClass={cs.CL},
      url={https://arxiv.org/abs/1904.09675}, 
}
